%% file: main_iclr.tex
\documentclass{article}
\usepackage{iclr2023_conference, times}[final]
\PassOptionsToPackage{numbers, compress}{natbib}
\usepackage{microtype}
\usepackage{graphicx}
\usepackage{subfigure}
\usepackage{booktabs} 

\usepackage{hyperref}

\usepackage{xcolor}
\definecolor{dark-red}{rgb}{0.4,0.15,0.15}
\definecolor{dark-blue}{rgb}{0.15,0.15,0.4}
\definecolor{medium-blue}{rgb}{0,0,0.5}
\hypersetup{
  colorlinks, linkcolor={dark-blue},
  citecolor={dark-blue}, urlcolor={medium-blue}
}

\usepackage{wrapfig}

\usepackage{xcolor}         

\usepackage{enumitem}
\usepackage{amsthm}
\newtheorem{theorem}{Theorem}

\usepackage[toc,page,header]{appendix}
\usepackage{minitoc}

\definecolor{bg}{gray}{0.95}

\input{math_commands}

\title{The Lie Derivative for Measuring Learned Equivariance}

\author{
Nate Gruver\thanks{Equal contribution.} , \, Marc Finzi$^*$, \, Micah Goldblum, \, Andrew Gordon Wilson \\
\, New York University \\
}

\iclrfinalcopy 
\begin{document}

\doparttoc 
\faketableofcontents 

\maketitle

\vspace{-2.5mm}

\begin{abstract}
Equivariance guarantees that a model's predictions capture key symmetries in data. When an image is translated or rotated, an equivariant model's representation of that image will translate or rotate accordingly. The success of convolutional neural networks has historically been tied to translation equivariance directly encoded in their architecture. The rising success of vision transformers, which have no explicit architectural bias towards equivariance, challenges this narrative and suggests that augmentations and training data might also play a significant role in their performance. In order to better understand the role of equivariance in recent vision models, we apply the Lie derivative, a method for measuring equivariance with strong mathematical foundations and minimal hyperparameters. Using the Lie derivative, we study the equivariance properties of hundreds of pretrained models, spanning CNNs, transformers, and Mixer architectures. The scale of our analysis allows us to separate the impact of architecture from other factors like model size or training method. Surprisingly, we find that many violations of equivariance can be linked to spatial aliasing in ubiquitous network layers, such as pointwise non-linearities, and that as models get larger and more accurate they tend to display more equivariance, regardless of architecture. For example, transformers can be more equivariant than convolutional neural networks after training.
\end{abstract}

\begin{figure}[h]
    \centering
    \begin{tabular}{ccc}
    \includegraphics[height=0.32\textwidth]{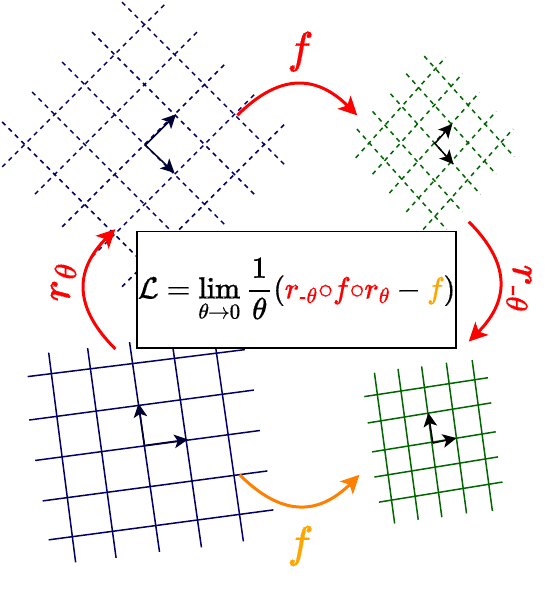} &
    \includegraphics[height=0.32\textwidth]{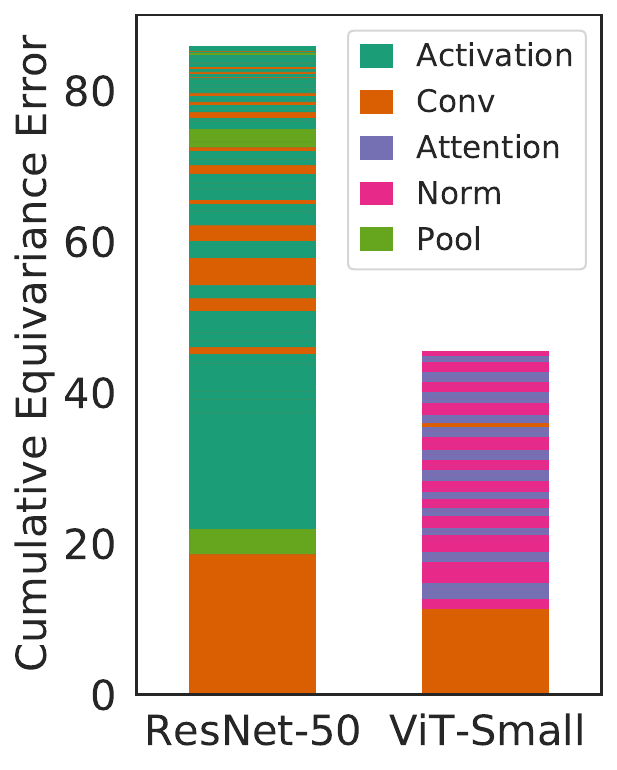} &
    \includegraphics[height=0.32\textwidth]{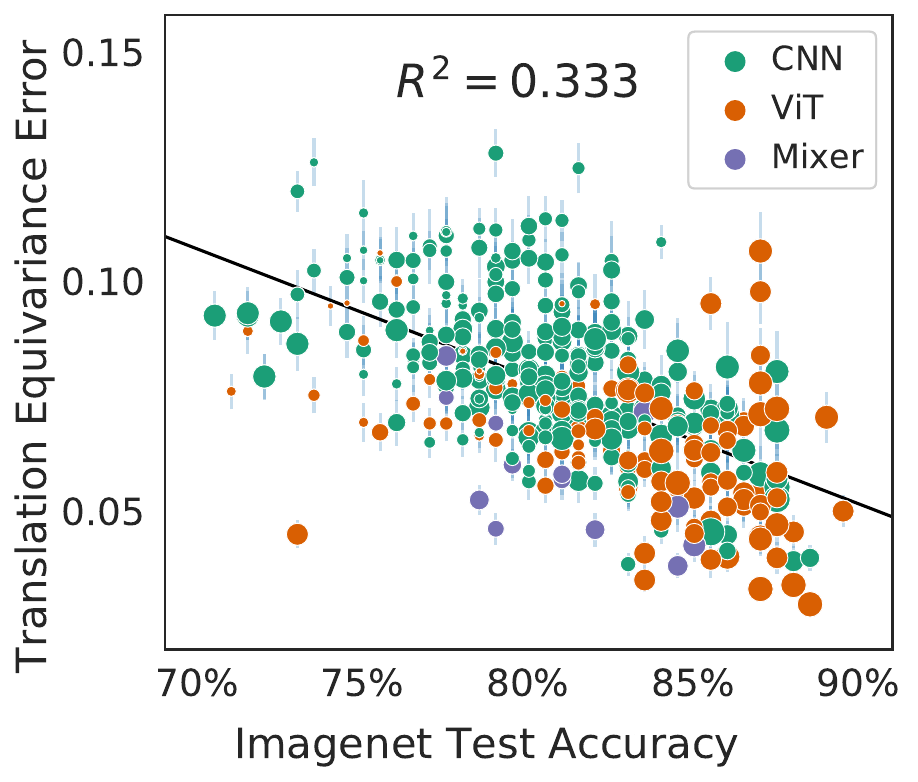} 
    \end{tabular}
    \vspace{-2mm}
    \caption{\textbf{(Left)}: The Lie derivative measures the equivariance of a function under continuous transformations, here rotation. \textbf{(Center)}: Using the Lie derivative, we quantify how much each layer contributes to the equivariance error of a model. Our analysis highlights surprisingly large contributions from non-linearities, which affects both CNNs and ViT architectures. \textbf{(Right)}: Translation equivariance as measured by the Lie derivative correlates with generalization in classification models, across convolutional and non-convolutional architectures. Although CNNs are often noted for their intrinsic translation equivariance, ViT and Mixer models are often more translation equivariant than CNN models after training.  
    }
    \vspace{-2mm}
    \label{fig:title-fig}
\end{figure}

\section{Introduction}
\label{sec:intro}

Symmetries allow machine learning models to generalize properties of one data point to the properties of an entire class of data points. A model that captures translational symmetry, for example, will have the same output for an image and a version of the same image shifted a half pixel 
to the left or right. If a classification model produces dramatically different predictions as a result of translation by half a pixel or rotation by a few degrees it is likely misaligned with physical reality. Equivariance provides a formal notion of consistency under transformation. A function is equivariant if symmetries in the input space are preserved in the output space.

Baking equivariance into models through architecture design has led to breakthrough performance across many data modalities, including images \citep{cohen2016group, veeling2018rotation}, proteins \citep{jumper2021highly} and atom force fields \citep{batzner20223, frey2022neural}. In computer vision, translation equivariance has historically been regarded as a particularly compelling property of convolutional neural networks (CNNs) \citep{lecun1995convolutional}. Imposing equivariance restricts the size of the hypothesis space, reducing the complexity of the learning problem and improving generalization \citep{goodfellow2016deep}. 

In most neural networks classifiers, however, true equivariance has been challenging to achieve, and many works have shown that model outputs can change dramatically for small changes in the input space \citep{azulay2018deep, engstrom2018rotation, vasconcelos2021impact, ribeiro2021convolutional}. Several authors have significantly improved the equivariance properties of CNNs with architectural changes inspired by careful signal processing \citep{zhang2019making, karras2021alias}, but non-architectural mechanisms for encouraging equivariance, such as data augmentations, continue to be necessary for good generalization performance \citep{wightman2021resnet}. 

The increased prominence of non-convolutional architectures, such as vision transformers (ViTs) and mixer models, simultaneously demonstrates that explicitly encoding architectural biases for equivariance is not necessary for good generalization in image classification, as ViT models perform on-par with or better than their convolutional counterparts with sufficient data and well-chosen augmentations \citep{dosovitskiy2020image,tolstikhin2021mlp}. Given the success of large flexible architectures and data augmentations, it is unclear what clear practical advantages are provided by explicit architectural constraints over learning equivariances from the data and augmentations. Resolving these questions systemically requires a unified equivariance metric and large-scale evaluation. 

In what follows, we introduce the Lie derivative as a tool for measuring the equivariance of neural networks under continuous transformations. The \emph{local equivariance error} (LEE), constructed with the Lie derivative, makes it possible to compare equivariance across models and to analyze the contribution of each layer of a model to its overall equivariance. Using LEE, we conduct a large-scale analysis of hundreds of image classification models. The breadth of this study allows us to uncover a novel connection between equivariance and model generalization, and the surprising result that ViTs are often more equivariant than their convolutional counterparts after training. To explain this result, we use the layer-wise decomposition of LEE to demonstrate how common building block layers shared across ViTs and CNNs, such as pointwise non-linearities, frequently give rise to aliasing and violations of equivariance. 

We make our code publicly available at
\href{https://github.com/ngruver/lie-deriv}{https://github.com/ngruver/lie-deriv}.

\section{Background}

\paragraph{Groups and equivariance}

Equivariance provides a formal notion of consistency under transformation. A function $f:V_1 \to V_2$ is equivariant to transformations from a symmetry group $G$ if applying the symmetry to the input of $f$ is the same as applying it to the output
\begin{equation}\label{eq:equivariance}
    \forall g\in G: \quad f(\rho_1(g)x) = \rho_2(g)f(x),
\end{equation}
where $\rho(g)$ is the \emph{representation} of the group element, which is a linear map $V\to V$.

The most common example of equivariance in deep learning is the translation equivariance of convolutional layers: if we translate the input image by an integer number of pixels in $x$ and $y$, the
output is also translated by the same amount, ignoring the
regions close to the boundary of the image. Here $x\in V_1=V_2$ is an image and the representation $\rho_1=\rho_2$ expresses translations of the image. The translation \emph{invariance} of certain neural networks is also an expression of the equivariance property, but where the output vector space $V_2$ has the trivial $\rho_2(g)=I$ representation, such that model outputs are unaffected by translations of the inputs. Equivariance is therefore a much richer framework, in which we can reason about  representations at the input and the output of a function. 

\paragraph{Continuous signals}

The inputs to classification models are discrete images sampled from a \textit{continuous} reality. Although discrete representations are necessary for computers, the goal of classification models should be learning functions that generalize in the real world. It is therefore useful to consider an image as a function $h: \R^2 \to \R^3$ rather than a discrete set of pixel values and broaden the symmetry groups we might consider, such as translations of an image by vector $\vb \in \R^2$, rather than an integer number of pixels. 

Fourier analysis is a powerful tool for understanding the relationship between continuous signals and discrete samples by way of frequency decompositions. Any $M \times M$ image, $h(\vx)$, can be constructed from its frequency components, $H_{nm}$, using a $2d$ Fourier series, $h(\vx)= \frac{1}{2\pi} \sum_{n,m} H_{nm} e^{2\pi i \vx \cdot [n,m]}$
where $\vx \in [0,1]^2$ and $n,m \in [\text{-}M/2,\text{-}M/2+1, ...,M/2]$, the bandlimit defined by the image size. 

\paragraph{Aliasing}
Aliasing occurs when sampling at a limited frequency $f_s$, for example the size of an image in pixels, causes high frequency content (above the Nyquist frequency $f_s/2$) to be converted into spurious low frequency content. Content with frequency $n$ is observed as a lower frequency contribution at frequency 
\begin{equation}
\mathrm{Alias}(n) = \left\{
\begin{array}{ll}
      n \bmod f_s &\text{if} \ (n \bmod f_s)<f_s/2 \\
      (n \bmod f_s) - f_s &\text{if} \ (n \bmod f_s)>f_s/2 \\
\end{array} 
\right\}.
\label{eq:alias}
\end{equation}
If a discretely sampled signal such as an image is assumed to have no frequency content higher than $f_s$, then the continuous signal can be uniquely reconstructed using the Fourier series and have a consistent continuous representation. But if the signal contains higher frequency content which gets aliased down by the sampling, then there is an ambiguity and exact reconstruction is not possible.

\paragraph{Aliasing and equivariance}
Aliasing is critically important to our study because it breaks equivariance to continuous transformations like translation and rotation. When a continuous image is translated its Fourier components pick up a phase:
$$h(\vx) \mapsto h(\vx - \vb) \implies H_{nm} \mapsto H_{nm} e^{-2\pi i \vb \cdot [n,m]}.$$
However, when an aliased signal is translated, the aliasing operation $\mathrm{A}$ introduces a scaling factor:
$$H_{nm} \mapsto H_{nm} e^{-2\pi i (b_0 \mathrm{Alias}(n) + b_1 \mathrm{Alias}(m))}$$
In other words, aliasing causes a translation \emph{by the wrong amount}: the frequency component $H_{nm}$ will effectively be translated by $[(\mathrm{Alias}(n)/n ) b_0, (\mathrm{Alias}(m) / m) b_1]$ which may point in a different direction than $\vb$, and potentially even the opposite direction. Applying shifts to an aliased image will yield the correct shifts for true frequencies less than the Nyquist but incorrect shifts for frequencies higher than the Nyquist. Other continuous transformations, like rotation, create similar asymmetries.  

Many common operations in CNNs can lead to aliasing in subtle ways, breaking equivariance in turn. \citet{zhang2019making}, for example, demonstrates that downsampling layers causes CNNs to have inconsistent outputs for translated images. The underlying cause of the invariance is aliasing, which occurs when downsampling alters the high frequency content of the network activations. The $M \times M$ activations at a given layer of a convolutional network have spatial Nyquist frequencies $f_s = M/2$. Downsampling halves the size of the activations and corresponding Nyquist frequencies. The result is aliasing of all nonzero content with $n \in [M/4,M/2]$. To prevent this aliasing, \citet{zhang2019making} uses a local low pass filter (Blur-Pool) to directly remove the problematic frequency regions from the spectrum. 

\begin{wrapfigure}{r}{0.4\textwidth}
\vspace{-6mm}
\includegraphics[width=0.4\textwidth]{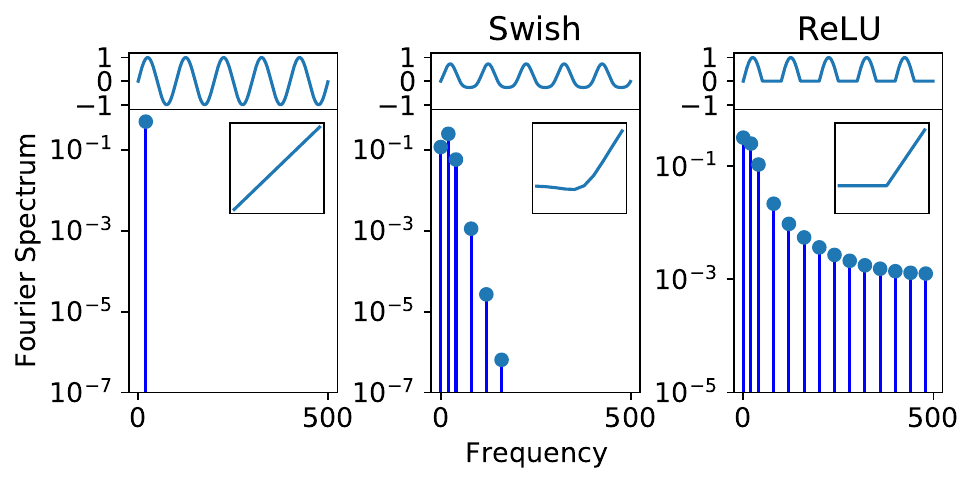}
\vspace{-8mm}
\caption{Non-linearities generate new high-frequency harmonics.}
\vspace{-4mm}
\label{fig:harmonics}
\end{wrapfigure}
While studying generative image models, \citet{karras2021alias} unearth a similar phenomenon in the 
pointwise nonlinearities of CNNs. Imagine an image at a single frequency $h(\vx) = \sin(2\pi \vx \cdot [n,m])$. Applying a nonlinear transformation to $h$ creates new high frequencies in the Fourier series, as illustrated in \autoref{fig:harmonics}. These high frequencies may fall outside of the bandlimit, leading to aliasing. To counteract this effect, \citet{karras2021alias} opt for smoother non-linearities and perform upsampling before calculating the activations. 

\vspace{-2mm}

\section{Related Work}
\label{sec:related-work}

While many papers propose architectural changes to improve the equivariance of CNNs \citep{zhang2019making, karras2021alias, e2cnn}, others focus purely on measuring and understanding how equivariance can emerge through learning from the training data \citep{lenc2015understanding}. \citet{olah2020naturally}, for example, studies learned equivariance in CNNs using model inversions techniques. While they uncover several fascinating properties, such as rotation equivariance that emerges naturally without architectural priors, their method is limited by requiring manual inspection of neuron activations. Most relevant to our work, \citet{bouchacourt2021grounding} measure equivariance in CNNs and ViTs by sampling group transformations. Parallel to our findings, they conclude that data and data augmentation play a larger role than architecture in the ultimate equivariance of a model. Because their study is limited to just a single ResNet and ViT architecture, however, they do not uncover the broader relationship between equivariance and generalization that we show here.

Many papers introduce consistency metrics based on sampling group transformations \citep{zhang2019making, karras2021alias, bouchacourt2021grounding}, but most come with significant drawbacks. When translations are sampled with an integer number of pixels \citep{zhang2019making, bouchacourt2021grounding}, aliasing effects will be completely overlooked. As a remedy, \citep{karras2021alias} propose a subpixel translation equivariance metric ($\text{EQ-T}_{\text{frac}}$) that appropriately captures aliasing effects. While this metric is a major improvement, it requires many design decisions not required by LEE, which has relatively few hyperparameters and seamlessly breaks down equivariance across layers. Relative to these other approaches, LEE offers a unifying perspective, with significant theoretical and practical benefits. 

\section{Measuring local equivariance error with Lie derivatives}
\label{sec:metric}

\paragraph{Lie Derivatives}

The Lie derivative gives a general way of evaluating the degree to which a function $f$ violates a symmetry. To define the Lie derivative, we first consider how a symmetry group element can transform a function by rearranging \autoref{eq:equivariance}:
\begin{equation*}
\rho_{21}(g)[f](x) = \rho_2(g)^{-1}f(\rho_1(g)x) \,.
\end{equation*}
The resulting linear operator, $\rho_{21}(g)[\cdot]$, acts on the vector space of functions, and $\rho_{21}(g)[f]=f$ if the function is equivariant. 
Every continuous symmetry group (Lie group), $G$, has a corresponding vector space (Lie algebra) $\mathfrak{g}=\mathrm{Span}(\{X_i\}_{i=1}^d)$,  with basis elements $X_i$ that can be interpreted as vector fields $\mathbb{R}^n \rightarrow \mathbb{R}^n$. For images, these vector fields encode infinitesimal transformations $\mathbb{R}^2 \rightarrow \mathbb{R}^2$ over the domain of continuous image signals $f: \mathbb{R}^2 \to \mathbb{R}^k$. One can represent group elements $g \in G$ (which lie in the connected component of the identity) as the \emph{flow} along a particular vector field $\Phi_Y^t$, where $Y=\sum_ia_iX_i$ is a linear combination of basis elements. The flow $\Phi_Y^t(x_0)$ of a point $x_0$ along a vector field $Y$ by value $t$ is defined as the solution to the ODE $\frac{dx}{dt} = Y(x)$ at time $t$ with initial value $x_0$. The flow $\Phi_Y^t$ smoothly parameterizes the group elements by $t$
so that the operator $\rho_{21}(\Phi_Y^t)[\cdot]$ connects changes in the space of group elements to changes in symmetry behavior of a function.

The Lie derivative of the function $f$ is the derivative of the operator $\rho_{21}(g)$ at $g=\mathrm{Identity} = \Phi_{0}$ along a particular vector field $Y$, 
\begin{equation}
    {\displaystyle {\mathcal {L}}_{Y}(f)} = \lim_{t\rightarrow 0} \frac{1}{t}(\rho_{21}(\Phi_{Y}^t)[f] - f) = \frac{d}{dt}\bigg|_{t=0} \rho_{21}(\Phi_{Y}^t)[f].
\end{equation}

Intuitively, the Lie derivative measures the sensitivity of a function to infinitesimal symmetry transformations. 
This local definition of equivariance error is related to the typical global notion of equivariance error. As we derive in Appendix
\ref{sec:local-global-lie-algebra}, if
$\forall i=1,...,d: {\mathcal {L}}_{X_i}(f)=0$ (and the exponential map is surjective)
then $\forall g\in G: \ f(\rho_1(g)x) = \rho_2(g)f(x)$ and for all $x$ in the domain, and vice versa. In practice, the Lie derivative is only a proxy for strict global equivariance. 
We note global equivariance includes radical transformations like translation by 75\% of an image, which is not necessarily desirable. In \autoref{sec:equiv-and-generalization} we show that our local formulation of the Lie derivative can capture the effects of many practically relevant transformations. 

\begin{figure*}[t!]
    \centering
    \includegraphics[width=\textwidth]{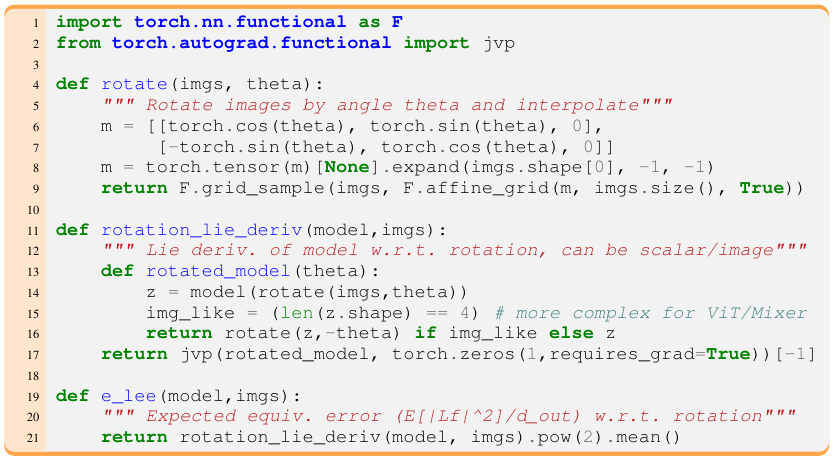}
    \vspace{-5mm}
    \caption{Lie derivatives can be computed using automatic differentiation. We show how a Lie derivative for continuous rotations can be implemented in PyTorch \citep{NEURIPS2019_9015}. The implementation in our experiments differs slightly, for computational efficiency and to pass second-order gradients through \textit{grid\_sample}.}
    \vspace{-2mm}
    \label{fig:lie-deriv-code}
\end{figure*}

The Lie derivative of a function with multiple outputs will also have multiple outputs, so if we want to summarize the equivariance error with a single number, we can compute the norm of the Lie derivative scaled by the size of the output. Taking the average of the Lie derivative over the data distribution, we define Local Equivariance Error (LEE),
\begin{equation} \label{eq:}
\mathrm{LEE}(f) = 
{\displaystyle \mathbb{E}_{x \sim \mathcal{D}} \|{\mathcal {L}}_{X}f(x)\|^2/\mathrm{dim}(V_2)}.
\end{equation} 
We provide a Python implementation of the Lie derivative calculation for rotations in \autoref{fig:lie-deriv-code} as an example. Mathematically, LEE also has an appealing connection to consistency regularization \citep{athiwaratkun2018there}, which we discuss in Appendix \ref{sec:consistency-reg}. 

\paragraph{Layerwise Decomposition of Lie Derivative}
Unlike alternative equivariance metrics, the Lie derivative decomposes naturally over the layers of a neural network, since it satisfies the chain rule.
As we show in Appendix \ref{app:chain}, the Lie derivative of the composition of two functions $h: V_1\rightarrow V_2$ and $f: V_2\rightarrow V_3$ satisfies
\begin{equation}
    {\displaystyle {\mathcal {L}}_{X}(f\circ h)}(x) = ({\mathcal {L}}_{X}f)(h(x)) + df|_{h(x)}({\mathcal {L}}_{X}h)(x),
    \label{eq:chain_rule}
\end{equation}
where $df|_{h(x)}$ is the Jacobian of $f$ at $h(x)$ which we will abbreviate as $df$. Note that this decomposition captures the fact that intermediate layers of the network may transform in their own way: 
\begin{align*}
f(h(x)) \mapsto &\rho_{31}(g)[f\circ g](x) = 
\rho_3(g)^{-1}f(\rho_2(g) \rho_2(g)^{-1} h(\rho_1(g)x)) =
\rho_{32}(g)[f] \circ \rho_{21}(g)[h](x)
\end{align*}
and the Lie derivatives split up accordingly.

Applying this property to an entire model as a composition of layers $\mathrm{NN}(x) = f_{N:1}(x):= f_N(f_{N-1}(...(f_1(x))))$, we can identify the contribution that each layer $f_i$ makes to the equivariance error of the whole. Unrolling \autoref{eq:chain_rule}, we have
\begin{equation}
    {\displaystyle {\mathcal {L}}_{X}(\mathrm{NN})} = \sum_{i=1}^N df_{N:i+1}{\mathcal {L}}_{X}f_i.
\end{equation}
Intuitively, the equivariance error of a sequence of layers is determined by the sum of the equivariance error for each layer multiplied by the degree to which that error is attenuated or magnified by the other layers (as measured by the Jacobian). We evaluate the norm of each of the contributions $df_{N:i+1}{\mathcal {L}}_{X}f_i$ to the (vector) equivariance error ${\displaystyle {\mathcal {L}}_{X}(\mathrm{NN})}$ which we compute using autograd and stochastic trace estimation, as we describe in Appendix \ref{app:estimator}. Importantly, the sum of norms may differ from the norm of the sum, but this analysis allows us to identify patterns across layers and pinpoint operations that contribute most to equivariance error. 

\section{Layerwise equivariance error}
\label{sec:equiv-and-architecture}

As described in \autoref{sec:related-work}, subtle architectural details often prevent models from being perfectly equivariant. Aliasing can result from careless downsampling or from an activation function with a wide spectrum. In this section, we explore how the Lie derivative uncovers these types of effects automatically, across several popular architectures. We evaluate the equivariance of pretrained models on 100 images from the ImageNet \citep{deng2009imagenet} test set.

\begin{figure*}[t!]
    \centering
    \setlength\tabcolsep{2pt}
    \begin{tabular}{cccc}
    \includegraphics[height=0.29\textwidth]{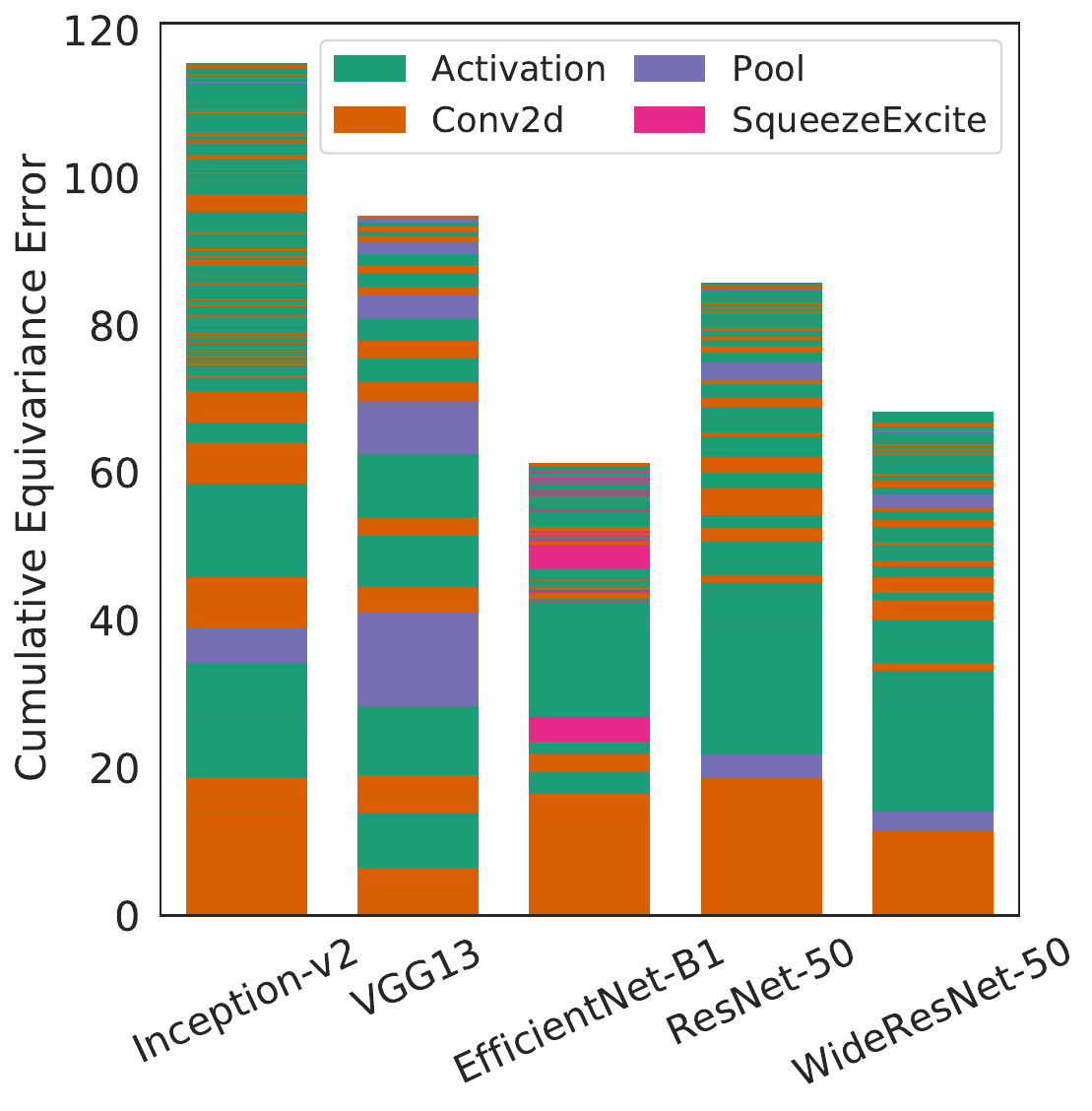} &  
    \includegraphics[height=0.29\textwidth]{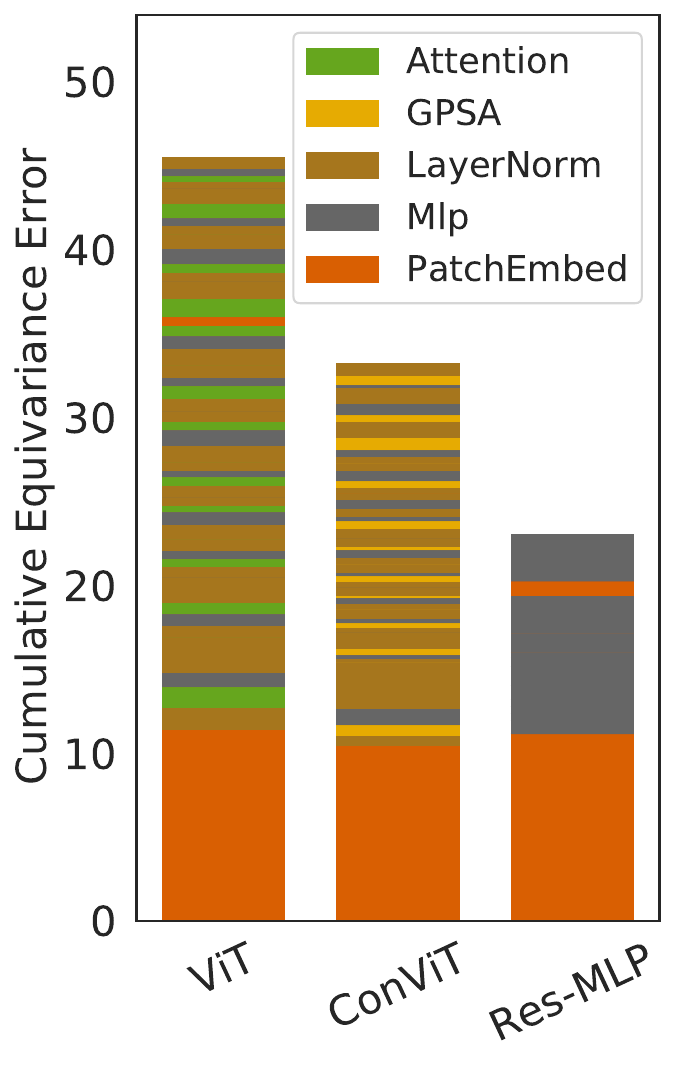} & 
    \includegraphics[height=0.29\textwidth]{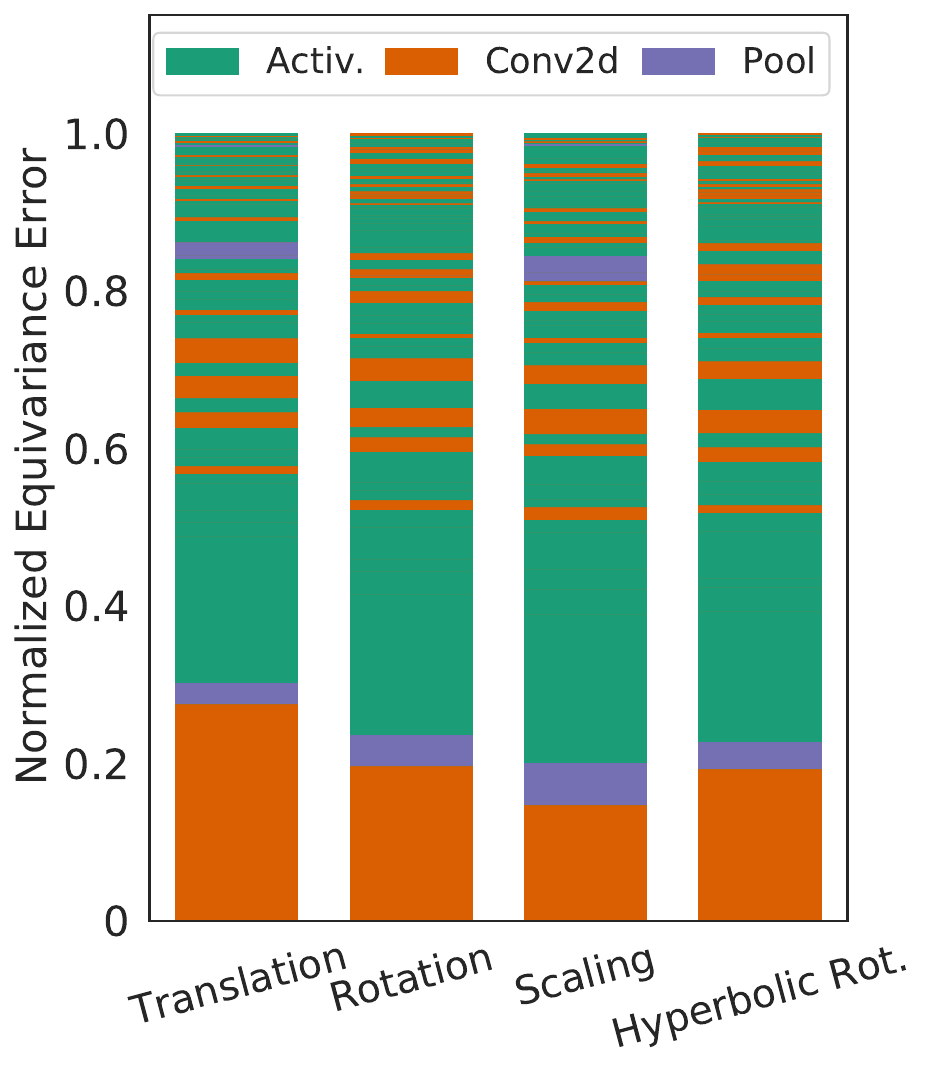} &
    \includegraphics[height=0.29\textwidth]{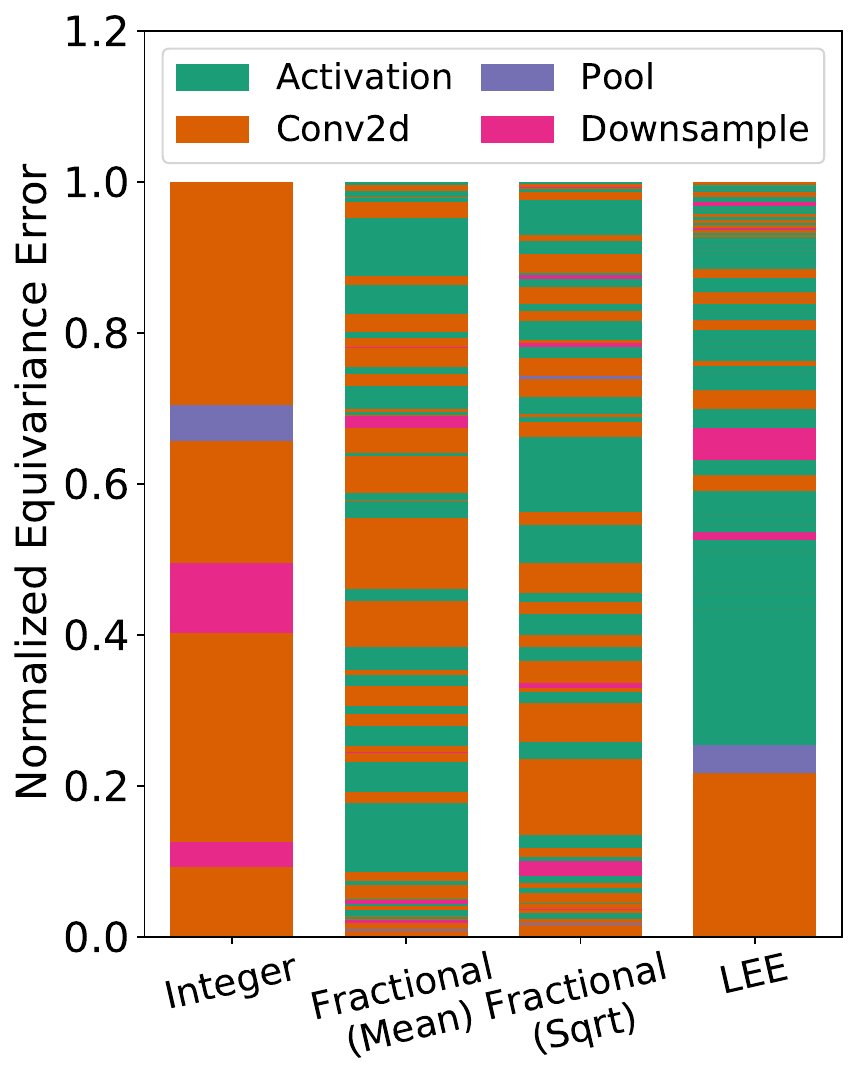} \\
    \end{tabular}
    \vspace{-3mm}
    \caption{Contributions to equivariance shown cumulatively by layer, in the order the layers occur in the network.
    \textbf{Left}: Convolutional architectures. In all the CNNs, much of the equivariance error comes from downsampling and non-linearities.
    \textbf{Middle-Left}: Non-convolutional architectures. The initial patch embedding, a strided convolution, is the largest contributor for the ViTs and Mixers. The rest of the error is distributed uniformly across other nonlinear operations. 
    \textbf{Middle-Right}: ResNet-50 across different transformations as a percentage. Despite being designed for translation equivariance, the fraction of equivariance error produced by each layer is almost identical for other affine transformations, suggesting that aliasing is the primary source of equivariance error.
    \textbf{Right}: Comparing LEE with alternative metrics for translation equivariance. Using integer translations misses key contributors to equivariance errors, such as activations, while using fractional translations can lead to radically different outcomes depending on choice of normalization ($N$ or $\sqrt{N}$). LEE captures aliasing effects and has minimal design decisions.
    }
    \vspace{-1mm}
    \label{fig:layerwise_architectures}
\end{figure*}

Using the layerwise analysis, we can dissect the sources of translation equivariance error in convolutional and non-convolutional networks as shown in \autoref{fig:layerwise_architectures} (left) and (middle-left). For the Vision Transformer and Mixer models, we see that the initial conversion from image to patches produces a significant portion of the error, and the remaining error is split uniformly between the other nonlinear layers: LayerNorm, tokenwise MLP, and self-attention. The contribution from these nonlinear layers is seldom recognized and potentially counterintuitive, until we fully grasp the deep connection between equivariance and aliasing. In \autoref{fig:layerwise_architectures} (middle-right), we show that this breakdown is strikingly similar for other image transformations like rotation, scaling, and hyperbolic rotations, providing evidence that the cause of equivariance error is not specific to translations but is instead a general culprit across a whole host of continuous transformations that can lead to aliasing. 

\setlength\tabcolsep{6pt}

We can make the relationship between aliasing and equivariance error precise by considering the aliasing operation $\mathrm{Alias}$ defined in \autoref{eq:alias}.
\begin{theorem}
For translations along the vector $v=[v_x,v_y]$, the aliasing operation $A$ introduces a translation equivariance error of
\begin{equation*}
    \|\mathcal{L}_v(A)(h)\|^2 = (2\pi)^2\sum_{n,m}H_{nm}^2 \big(v_x^2(\mathrm{Alias}(n)-n)^2+v_y^2(\mathrm{Alias}(m)-m)^2\big),
\end{equation*}
where $h(\vx) =\tfrac{1}{2\pi}\sum_{n,m}H_{nm}e^{2\pi i \vx \cdot [n,m]}$ is the Fourier series for the input image $h$.
\end{theorem}
We provide the proof in Appendix \ref{sec:trans_aliasing_equivariance}.
The connection between aliasing and LEE is important because aliasing is often challenging to identify despite being ubiquitous \citep{zhang2019making, karras2021alias}. Aliasing in non-linear layers impacts all vision models and is thus a key factor in any fair comparison of equivariance. 

As alternative equivariance metrics exist, it is natural to wonder whether they can also be used for layerwise analysis. In \autoref{fig:layerwise_architectures} (right), we show how two equivariance metrics from \citet{karras2021alias} compare with LEE, highlighting notable drawbacks. (1) Integer translation equivariance completely ignores aliasing effects, which are captured by both LEE and fractional translations. (2) Though fractional translation metrics ($\text{EQ-T}_{\text{frac}}$) correctly capture aliasing, comparing the equivariance of layers with different resolutions ($C \times H \times W$) requires an arbitrary choice of normalization. This choice can lead to radically different outcomes in the perceived contribution of each layer and is not required when using LEE, which decomposes across layers as described in \autoref{sec:metric}. We provide a detailed description of the baselines in Appendix \ref{sec:layerwise-equivariance-baselines}. 

\section{Trends in learned equivariance}
\label{sec:equiv-and-generalization}

\paragraph{Methodology}

We evaluate the Lie derivative of many popular classification models under transformations including $2d$ translation, rotation, and shearing. We define continuous transformations on images using bilinear interpolation with reflection padding. In total, we evaluate 410 classification models, a collection comprising popular CNNs, vision transformers, and MLP-based architectures \citep{rw2019timm}. Beyond diversity in architectures, there is also substantial diversity in model size, training recipe, and the amount of data used for training or pretraining. This collection of models therefore covers many of the relevant axes of variance one is likely to consider in designing a system for classification. We include an exhaustive list of models in the Appendix \ref{sec:model-list}.

\begin{figure}[t!]
    \setlength\tabcolsep{0pt}
    \begin{tabular}{ccc}
    \centering
    \includegraphics[height=0.14\textheight]{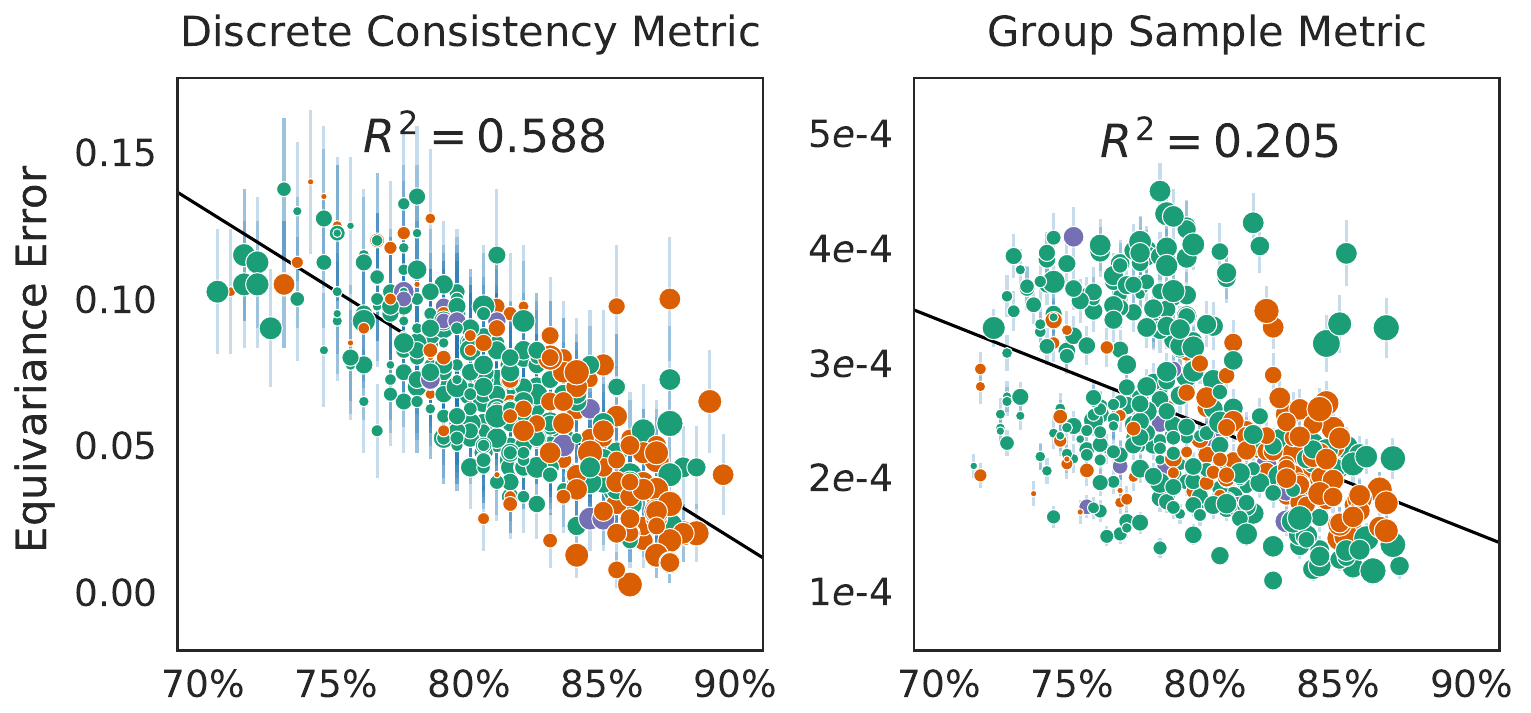} & \hspace{5mm} &
    \includegraphics[height=0.14\textheight]{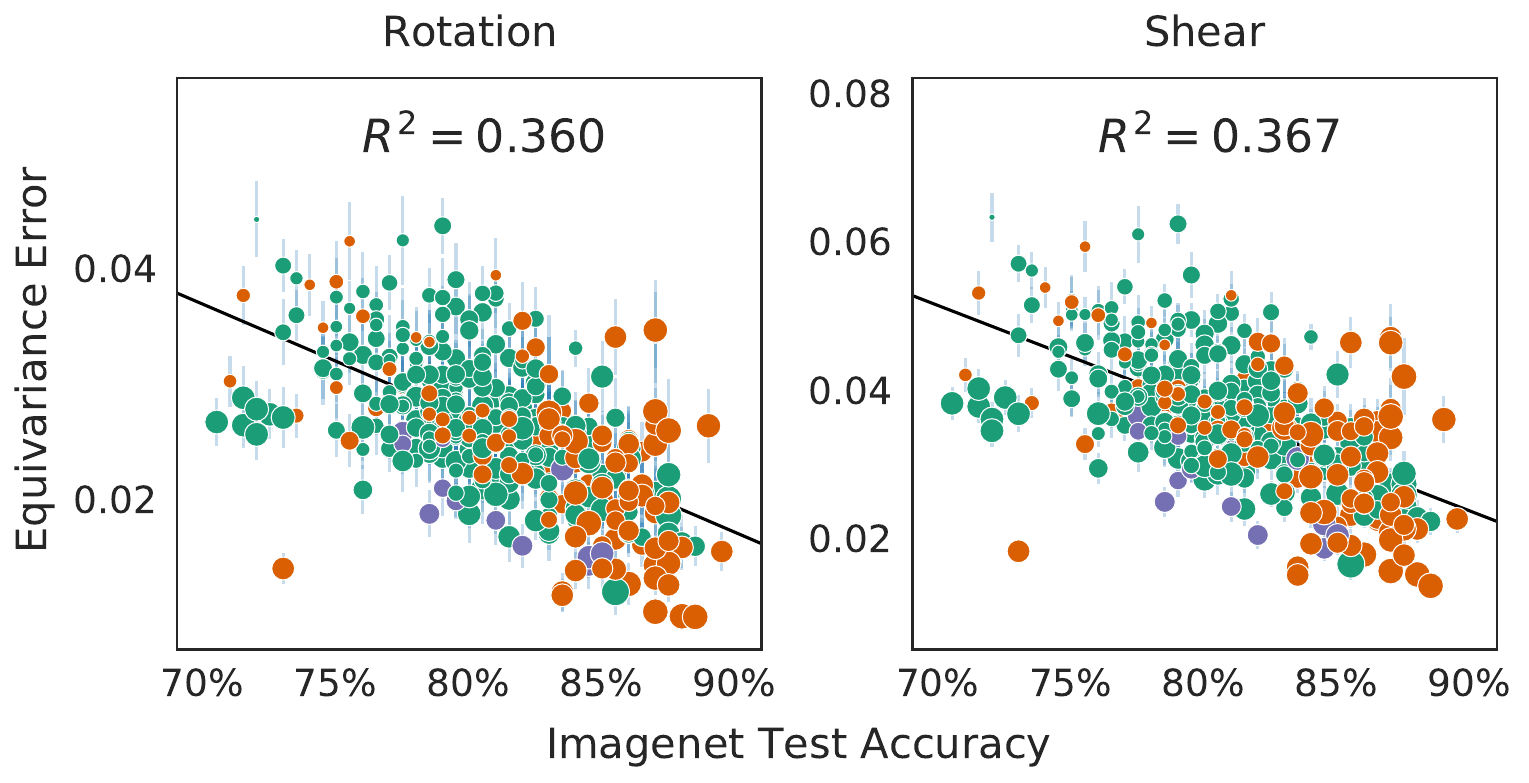}
    \end{tabular}
    \includegraphics[height=0.03\textwidth]{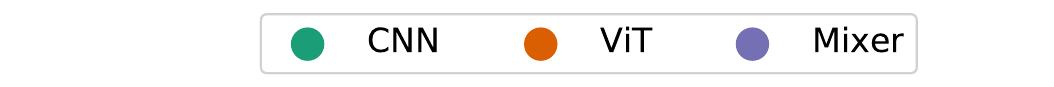}
    \includegraphics[height=0.03\textwidth]{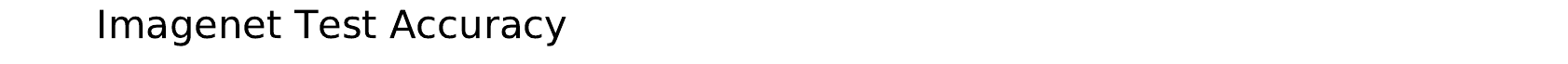}
    \setlength\tabcolsep{6pt}
    \vspace{-6mm}
    \caption{
    Equivariance metrics evaluated on the ImageNet test set.
    \textbf{Left:}
    Non-LEE equivariance metrics display similar trends to \autoref{fig:title-fig}, despite using larger, multi-pixel transformations.
    \textbf{Right:}
    Norm of rotation and shear Lie derivatives. Across all architectures, models with strong generalization become more equivariant to many common affine transformations. Marker size indicates model size. Error bars show one standard error over test set images used in the equivariance calculation.
    }
    \label{fig:lie_deriv}
\end{figure}

\paragraph{Equivariance across architectures} 

As shown in \autoref{fig:title-fig} (right), the translation equivariance error (Lie derivative norm) is strongly correlated with the ultimate test accuracy that the model achieves. Surprisingly, despite convolutional architectures being motivated and designed for their translation equivariance, we find no significant difference in the equivariance achieved by convolutional architectures and the equivariance of their more flexible ViT and Mixer counterparts when conditioning on test accuracy. This trend also extends to rotation and shearing transformations, which are common in data augmentation pipelines \citep{cubuk2020randaugment} (in \autoref{fig:lie_deriv} (right)). Additional transformation results included in Appendix \ref{sec:more-transforms}.

For comparison, we also evaluate the same set of models using two alternative equivariance metrics: prediction consistency under discrete translation \citep{zhang2019making} and expected equivariance under group samples \citep{finzi2020generalizing,hutchinson2021lietransformer}, which is similar in spirit to $\text{EQ-T}_{\text{frac}}$ \citep{karras2021alias}
(exact calculations in Appendix \ref{sec:alt-equivariance-metrics}). Crucially, these metrics are slightly less \emph{local} than LEE, as they evaluate equivariance under transformations of up to 10 pixels at a time. The fact that we obtain similar trends highlights LEE's relevance beyond subpixel transformations. 

\paragraph{Effects of Training and Scale}

In \autoref{sec:related-work} we described many architectural design choices that have been used to improve the equivariance of vision models, for example \citet{zhang2019making}'s Blur-Pool low-pass filter. We now investigate how equivariance error can be reduced with non-architectural design decisions, such as increasing model size, dataset size, or training method. Surprisingly, we show that equivariance error can often be significantly reduced without any changes in architecture.

\begin{figure*}[t!]
    \centering

    \includegraphics[width=\textwidth]{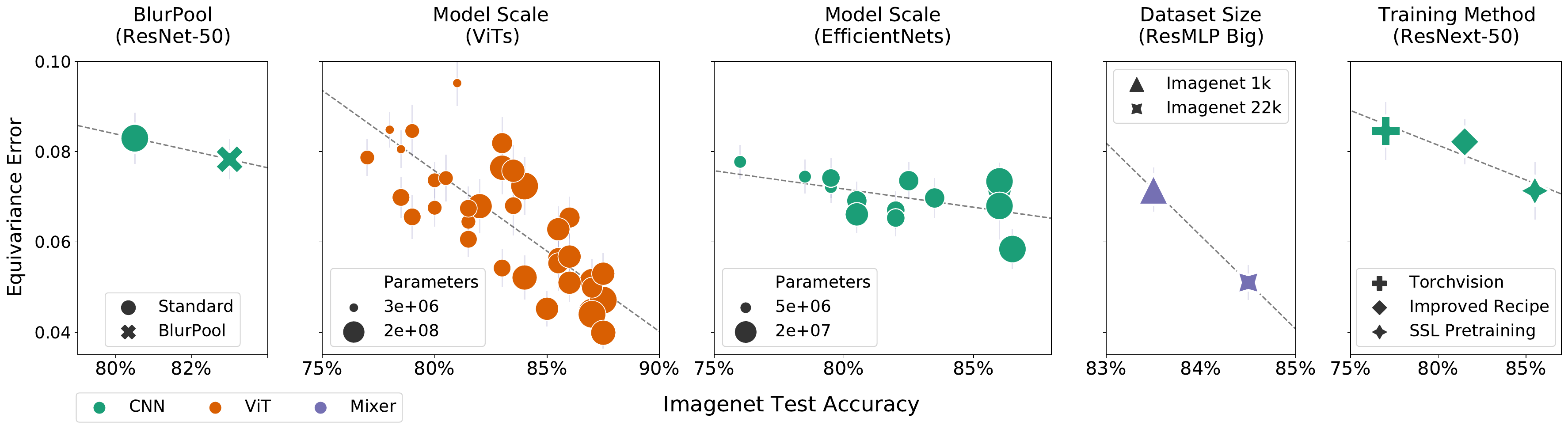} 
    \vspace{-4mm}
    \caption{Case studies in decreasing translational equivariance error, numbered left-to-right. \textbf{1}: Blur-Pool  \citep{zhang2019making}, an architectural change to improve equivariance, decreases the equivariance error but by less than can be accomplished by improving the training recipe or increasing the scale of model or dataset. \textbf{2-3}: Increasing the number of parameters for a fixed model family (here ViTs \citep{el2021xcit} and EfficientNets \citep{tan2019efficientnet}). \textbf{4}: Increasing the training dataset size for a ResMLP Big \citep{touvron2021resmlp} model. \textbf{5}: Changing the training recipe for ResNeXt-50 \citep{xie2017aggregated} with improved augmentations \citep{wightman2021resnet} or SSL pretraining \citep{DBLP:journals/corr/abs-1905-00546}.   Error bars show one standard error over images in the Lie derivative calculation.}
    \vspace{-2mm}
    \label{fig:scale-comparison}
\end{figure*}

In \autoref{fig:scale-comparison}, we show slices of the data from \autoref{fig:title-fig} along a shared axis for equivariance error. As a point of comparison, in \autoref{fig:scale-comparison} (left), we show the impact of the Blur-Pool operation discussed above on a ResNet-50 \citep{zhang2019making}. In the accompanying four plots, we show the effects of increasing model scale (for both ViTs and CNNs), increasing dataset size, and finally different training procedures. Although \citet{zhang2019making}'s architectural adjustment does have a noticeable effect, factors such as dataset size, model scale, and use of modern training methods, have a much greater impact on learned equivariance.

As a prime example, in \autoref{fig:scale-comparison} (right), we show a comparison of three training strategies for ResNeXt-50 -- an architecture almost identical to ResNet-50. We use \citet{wightman2021resnet}'s pretrained model to illustrate the role of an improved training recipe and \citet{wslimageseccv2018}'s semi-supervised model as an example of scaling training data. Notably, for a fixed architecture and model size, these changes lead to decreases in equivariance error on par with architectural interventions (BlurPool). This result is surprising when we consider that \citet{wightman2021resnet}'s improved training recipe benefits significantly from Mixup \citep{zhang2017mixup} and CutMix \citep{yun2019cutmix}, which have no obvious connection to equivariance. 
Similarly, \citet{wslimageseccv2018}'s semi-supervised method has no explicit incentive for equivariance.

\paragraph{Equivariance out of distribution}

\begin{figure*}[t!]
    \centering
    \includegraphics[width=\textwidth]{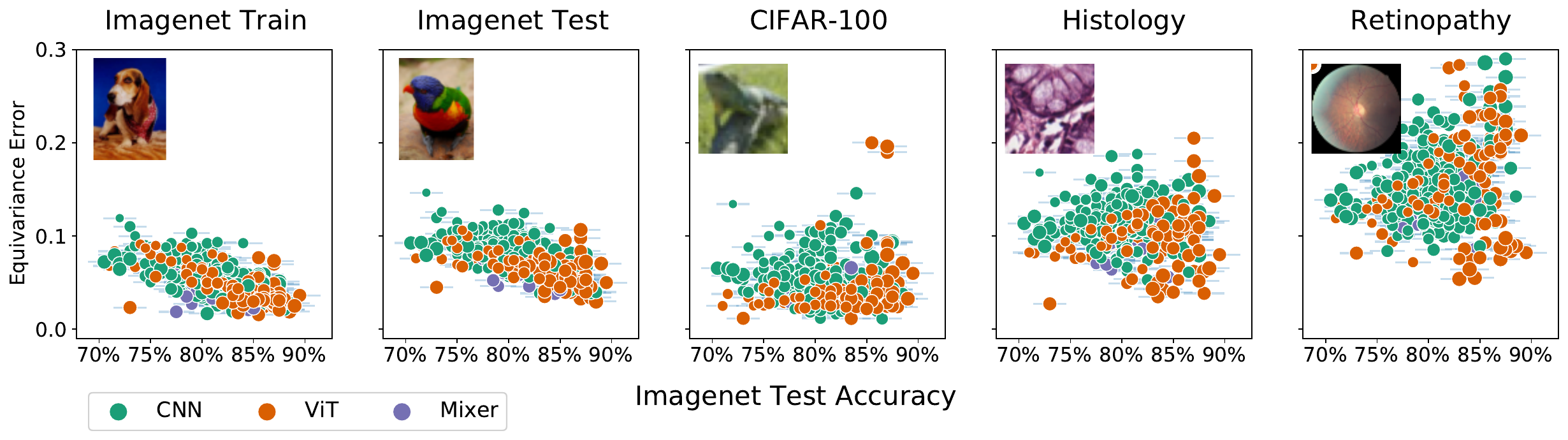}
    \vspace{-5mm}
    \caption{Models are less equivariant on test data and becoming decreasingly equivariant as the data moves away from the training manifold. As examples of data with similar distributions, we show equivariance error on the ImageNet train and test sets as well as CIFAR-100. As examples of out-of-distribution data, we use two medical datasets (which often use Imagenet pretraining), one for Histology \citep{kather2016multi} and one for Retinopathy \citep{kaggle-diabetic-retinopathy}.}
    \label{fig:equivariance_gap}
\end{figure*}

From our analysis above, large models appear to learn equivariances that rival architecture engineering in the classification setting. When learning equivariances through data augmentation, however, there is no guarantee that the equivariance will generalize to data that is far from the training distribution. Indeed, \citet{engstrom2019exploring} shows that carefully chosen translations or rotations can be as devastating to model performance as adversarial examples. We find that vision models do indeed have an \emph{equivariance gap}: models are less equivariant on test data than train, and this gap grows for OOD inputs as shown in Figure \ref{fig:equivariance_gap}.
Notably, however, architectural biases do not have a strong effect on the equivariance gap, as both CNN and ViT models have comparable gaps for OOD inputs. 

\paragraph{Why aren't CNNs more equivariant than ViTs?}

Given the deep historical connection between CNNs and equivariance, the results in \autoref{fig:lie_deriv} and \autoref{fig:equivariance_gap} might appear counterintuitive. ViTs, CNNs, and Mixer have quite different inductive biases and therefore often learn very different representations of data \citep{raghu2021vision}. Despite their differences, however, all of these architectures are fundamentally constructed from similar building blocks--such as convolutions, normalization layers, and non-linear activations which can all contribute to aliasing and equivariance error. Given this shared foundation, vision models with high capacity and effective training recipes are more capable of fitting equivariances already present in augmented training data. 

\paragraph{Learning rotation equivariance}

\begin{wraptable}{r}{0.39\textwidth}
  \setcitestyle{numbers}
  \setcitestyle{square}
  \vspace{1mm}
  \begin{tabular}{c|c}
  Model & Test Error (\%) \\
  \hline
  G-CNN \citep{cohen2016group} & 2.28 \\
  H-NET \citep{worrall2017harmonic} & 1.69\\
  ORN \citep{zhou2017oriented} & 1.54 \\
  TI-Pooling \citep{laptev2016ti} & 1.2\\
  Finetuned MAE & 1.14 \\
  RotEqNet \citep{marcos2017rotation} & 1.09 \\
  E(2)-CNN \citep{e2cnn} & 0.68 \\ 
  \end{tabular}
  \setcitestyle{authoryear}
  \setcitestyle{round}
  \caption{Our finetuned MAE is competitive with several architectures explicitly engineered to encode rotation invariance on RotMNIST, where rotation invariance is clearly crucial to generalization.}
  \vspace{-2mm}
  \label{tab:rotmnist}
\end{wraptable}
We finally consider the extent to which large-scale pretraining can match strong architectural priors in a case where equivariance is obviously desirable. We fine-tune a state-of-the-art vision transformer model pretrained with masked autoencoding \citep{MaskedAutoencoders2021} for 100 epochs on rotated MNIST \citep{e2cnn} (details in Appendix \ref{sec:mae-finetune}). This dataset, which contains MNIST digits rotated uniformly between -180 and 180 degrees, is a common benchmark for papers that design equivariant architectures. In \autoref{tab:rotmnist} we show the test errors for many popular architectures with strict equivariance constrainets alongside the error for our finetuned model. Surprisingly, the finetuned model achieves competitive test accuracy, in this case a strong proxy for rotation invariance. Despite having relatively weak architectural biases, transformers are capable of learning and generalizing on well on symmetric data. 

\section{Conclusion}

We introduced a new metric for measuring equivariance which enables a nuanced investigation of how architecture design and training procedures affect representations discovered by neural networks. Using this metric we are able to pinpoint equivariance violation to individual layers, finding that pointwise nonlinearities contribute substantially even in networks that have been designed for equivariance. We argue that aliasing is the primary mechanism for how equivariance to continuous transformations are broken, which we support theoretically and empirically. We use our measure to study equivariance learned from data and augmentations, showing model scale, data scale, or training recipe can have a greater effect on the ability to learn equivariances than architecture.

Many of these results are contrary to the conventional wisdom. For example, transformers can be more equivariant than convolutional neural networks after training, and can learn equivariances needed to match the performance of specially designed architectures on benchmarks like rotated MNIST, despite a lack of explicit architectural constraints. These results suggest we can be more judicious in deciding when explicit interventions for equivariance are required, especially in many real world problems where we only desire approximate and local equivariances.On the other hand, explicit constraints will continue to have immense value when exact equivariances and extrapolation are required --- such as rotation invariance for molecules. Moreover, despite the ability to learn equivariances on training data, we find that there is an equivariance \emph{gap} on test and OOD data which persists regardless of the model class. Thus other ways of combating aliasing outside of architectural interventions may be the path forward for improving the equivariance and invariance properties of models. 

\clearpage

\clearpage
\bibliography{references}
\bibliographystyle{iclr2023_conference}
\clearpage

\appendix
\onecolumn

\addcontentsline{toc}{section}{Appendix} 
\part{Appendix}
\parttoc

\section{Aliasing Extended Discussion}
\label{sec:signal-processing-background}

When working with signal $f$ we can write the fourier series
$$f(k) = \sum_{n=-\infty}^{\infty} F(n) e^{i 2\pi k n / N}$$
where $F(n)$ are the frequency components and the values $n$ are called \textit{harmonics}.
Take the $f$ to be discretely sampled at uniformly spaced points in a bounded interval $[-1, -1 + \Delta, ..., 1-\Delta, 1]$, with $\Delta = 1 / K$. Because the sampling rate is limited, it is impossible to correctly measure components $F(n)$ where $n > K / 2$. Evaluated only at the grid points, such content could have identical values to components with lower frequencies, causing fundamental ambiguities:
$$
\sin(2\pi(k + nK) t + \phi) = 
\begin{cases}
+ \sin(2\pi(k + nK) t + \phi) & k + nK \geq 0 \\
- \sin(2\pi|k + nK| t + \phi) & k + nK < 0
\end{cases}
$$
The default mechanism for resolving these ambiguities in the reconstruction is to choose the lowest frequency component for the corresponding observations, leading to the aliasing operation given in \autoref{eq:alias}. 

This operation can also be considered a translation in the frequency domain. Crucially, operations in frequency domain have corresponding operations in the spatial domain, and thus aliasing can give rise to recognizable patterns in images with poorly chosen resolutions, for example moire patterns. This relationship also means aliasing's effects on translations in frequency space, for example, can effect translational spatial symmetries. 

To make this relationship explicit, let us consider the translational symmetries of the first set of feature maps in a CNN in two scenarios. In both scenarios the transformation is downward translation of the input by 10\% of its height. First, let us consider the case where this transformation happens to result in a translation by a discrete numbers of pixels, $p$ in the feature maps. Obviously the pixels at the bottom of the image become lost to the boundary and thus cannot be recovered from the corresponding feature maps, as would be required for equivariance, as illustrated in \autoref{fig:title-fig}. As the amount of this translation gets smaller and smaller, however, the effect of the boundary should decrease, and yet the ability to recover the image can still be strongly affected by innate signal processing properties. 

Consider the case where the CNN has a stride of 2. The feature maps will have half the width of the original image. Therefore the Nyquist frequency will also be half that of the Nyquist frequency of the image, and there will be aliasing of all the frequencies in between the original Nyquist frequency and the new value. When we try to reverse the transformation by translating $p$ pixels upwards, the resulting translation will no longer be the inverse of the translation on the image. Therefore we cannot achieve perfect equivariance. 

As another important subcase, let's also consider the non-linear activation in the CNN layer by itself. If we apply the non-linearity to a translated input, we can simply use the fact the result was a discrete translation in the output space to map the values at the grid points to values at different grid points under the reverse transformation. In this case there is clearly no issue introduced from the frequency domain properties of non-linearities on their own. 

Now let's consider a translation of $1/p$ pixels. In this case, reconstructing the image after the translation is non-trivial, and we need to perform interpolation to calculate the values of the corresponding continuous image at the points that will become translated to the evaluation points. In order to perform this interpolation we must actually consider the full frequency spectrum of the image. Now the effects of pointwise non-linearities can become apparent. Because non-linearities can introduce high frequency content, these high frequencies become important when reconstructing the signal using interpolation. Aliasing makes this reconstruction fundamentally challenging and thus equivariance is impossible to achieve. 

\section{Lie Groups, Lie Derivatives, and LEE}

\subsection{Lie Groups and Local/Global Notions of Equivariance}
\label{sec:local-global-lie-algebra}

The key to understanding why the local - global equivalence holds is that $(\exp(X)-1) = \sum_{k=1}^\infty X^k/k!$ has the same nullspace as $X$ (here repeated application of $X$ on a function $f$ is just the repeated directional derivative, and this is the definition of a vector field used in differential geometry). Since they have the same nullspace, the space of functions for which $\exp(X)f=f$ is the same as the space $Xf=0$. The same principle holds for $\rho(\exp(X))f=f$ and $d\rho(X)f=0$ since $\rho(\exp(X)) = \exp(d\rho(X))$ (a basic result in representation theory, which can be found in \citep{hall2013lie}) where $d\rho$ is the corresponding Lie algebra representation of $\rho$, which for vector fields is the Lie derivative $d\rho(X)=L_X$. Hence carrying over the constraint for each element $\forall X \in \mathfrak{g}: L_Xf=0$ is equivalent to $\forall X\in \mathfrak{g}: \rho(\exp(X))f=f$ which is the same as $\forall g\in G: \rho(g)f=f$. Unpacking the representation $\rho_{12}$ of $f$, this is just the global equivariance constraint $\forall g\in G: \rho_2(g)^{-1}f(\rho_1(g)x)=f(x)$.

\subsection{Lie Derivative Chain Rule}\label{app:chain}
Suppose we have two functions $h: V_1\rightarrow V_2$ and $f: V_2\rightarrow V_3$, and corresponding representations $\rho_1,\rho_2,\rho_3$ for the vector spaces $V_1,V_2,V_3$. Expanding out the definition of $\rho_{31}$,
\begin{align*}
    \rho_{31}(g)[f\circ h](x) &= \rho_3(g)^{-1}f(h(\rho_1(g)x))\\
    &=\rho_3(g)^{-1}f(\rho_2(g) \rho_2(g)^{-1} h(\rho_1(g)x)) \\
    &= \rho_{32}(g)[f] \circ \rho_{21}(g)[h](x).
\end{align*}

From the definition of the Lie derivative, and using the chain rule that holds for the derivative with respect to the scalar $t$, and noting that $g_0=\mathrm{Id}$ so $\rho(g_0)=\mathrm{Id}$, we have
\begin{align*}
    {\displaystyle {\mathcal {L}}_{X}(f\circ h)}(x) &=\frac{d}{dt}\bigg( \rho_{31}(g_t)[f\circ h](x)\bigg)\bigg|_0 \\
    &= \frac{d}{dt}\bigg(\rho_{32}(g_t)[f] \circ \rho_{21}(g_t)[h](x)\bigg)\bigg|_0\\
    &= \bigg(\frac{d}{dt}\rho_{32}(g_t)[f]\big|_{t=0}\bigg) \circ \rho_{21}(g_0)[h](x) + \bigg[d(\rho_{32}(g_0)[f])\bigg|_{h(x)}\bigg] \bigg(\frac{d}{dt}\rho_{21}(g_t)[h]\big|_{t=0}\bigg)(x)\\
    &= \bigg(\frac{d}{dt}\rho_{32}(g_t)[f]\big|_{t=0}\bigg) \circ h(x) + df|_{h(x)}\bigg(\frac{d}{dt}\rho_{21}(g_t)[h]\big|_{t=0}\bigg)(x)\\
    &= ({\mathcal {L}}_{X}f)\circ h(x) + df|_{h(x)}({\mathcal {L}}_{X}h)(x),
\end{align*}
where $df|_{h(x)}$ is the Jacobian of $f$ at $h(x)$ and $df|_{h(x)}({\mathcal {L}}_{X}h)(x)$ is understood to be the Jacobian vector product of $df|_{h(x)}$ with $({\mathcal {L}}_{X}h)(x)$, equivalent to the directional derivative of $f$ along $({\mathcal {L}}_{X}h)(x)$. Therefore the Lie derivative satisfies a chain rule

\subsection{Stochastic Trace Estimator for Layerwise Metric}\label{app:estimator}

Unrolling this chain rule for a sequence of layers $\mathrm{NN}(x) = f_{N:1}(x):= f_N(f_{N-1}(...(f_1(x))))$, or even an autograd DAG, we can identify the contribution that each layer $f_i$ makes to the equivariance error of the whole as the sum of terms $C_i = df_{N:i+1}{\mathcal {L}}_{X}f_i$,
   ${\displaystyle {\mathcal {L}}_{X}(\mathrm{NN})} = \sum_{i=1}^N C_i.$
   
Each of these $C_i$, like ${\displaystyle {\mathcal {L}}_{X}(\mathrm{NN})}$ measure the equivariance error for all of the outputs (which we define to be the softmax probabilities), and are hence vectors of size $K$ where $K$ is the number of classes. In order to summarize the $C_i$ as a single number for plotting, we compute their norm $\|C_i\|$ which satisfy $\|{\displaystyle {\mathcal {L}}_{X}(\mathrm{NN})}\| \le \sum_i \|C_i\|$.

To compute $df_{N:i+1}{\mathcal {L}}_{X}f_i$, one can use autograd to perform Jacobian vector products (as opposed to typical vector Jacobian products) and build up $df_{N:i+1}$ in a backwards pass. Unfortunately doing so is quite cumbersome in the PyTorch framework where the large number of available models are implemented and pretrained. A trick which can be used to speed up this computation is to use stochastic trace estimation \citep{avron2011randomized}.
Since vector Jacobian products are cheap and easy, we can compute $\|C_i\|^2 = \mathbb{E}[\hat{A}]$  as the expectation of the estimator $\hat{A} = (1/N)\sum_{n}^N (z_n^\top C_i)^2 = (1/N)\sum_{n}^N(z_n^\top df_{N:i+1}{\mathcal {L}}_{X}f_i)^2$ with iid. Normal probe vectors $z_n\sim \mathcal{N}(0,I)$, and the quantity $z_n^\top df_{N:i+1}$ which is a standard vector Jacobian product. 

One can see that $\mathbb{E}[\hat{A}] = C_i^\top \mathbb{E}[zz^\top]C_i = C_i^\top I C_i = \|C_i\|^2$. We can then measure the variance of this estimator to control for the error and increase $N$ until this error is at an acceptable tolerance (we use $N=100$ probes). The convergence of this trace estimator is shown in \autoref{fig:shear-hyperbolic} (right) for several different layers of a ResNet-50. In producing the final layerwise attribution plots, we average the computed quantity $\|C_i\|$ over $20$ images from the ImageNet test set.

\section{LEE Theorems}

\subsection{LEE and consistency regularization}
\label{sec:consistency-reg}

As shown in \citet{athiwaratkun2018there}, consistency regularization with Gaussian input perturbations can be viewed as an estimator for the norm of the Jacobian of the network, but in fact when the perturbations are not Gaussian but from small spatial transformations, consistency regularization actually penalizes the Lie derivative norm. In the $\Pi$-model \citep{laine2016temporal} (the most basic form of consistency regularization), the consistency regularization minimizes the norm of the difference of the outputs of the network when two randomly sampled transformations $T^a$ and $T^b$ are applied to the input,
\begin{equation}
    \mathrm{L}_{\mathrm{cons}} = \|f(T^a(x))-f(T^b(x))\|^2.
\end{equation}
Suppose that the two transformations are representations of a given symmetry group and can be written as $T^a = \rho(g_a)$ and $T^b = \rho(g_b)$, and the group elements can be expressed as the flow generated by a linear combination of the vector fields which form the Lie Algebra: $g_a = \Phi_{\sum_i a_i X_i}$ for some coefficients $\{a_i\}_{i=1}^d$ and likewise for $g_b$. We can define the log map, mapping group elements top their generator values in this basis: $\log(g_a)=a$. Then, assuming $a_i$ are small (and therefore the transformations are small), Taylor expansion yields
$\mathrm{L}_{\mathrm{cons}} =\|f(x)+\sum_i a_i\mathcal{L}_{X_i}f(x) +O(a^2) -[f(x) + \sum_j b_j\mathcal{L}_{X_j}f(x) + O(b^2)]\|^2$. Therefore, taking the expectation over the distribution which $a$ and $b$ are sampled over (which is assumed to be centered with $\mathbb{E}[a_i]=\mathbb{E}[b_i]=0$ as well as the input distribution $x$, we get that
\begin{equation}
    \mathbb{E}_{a,b,x}[\mathrm{L}_{\mathrm{cons}}] = 2\mathbb{E}[\|\sum_i\mathcal{L}_{X_i}f(x)\|^2_\Sigma] + \text{higher order terms},
\end{equation}
where $\|\|^2_\Sigma$ denotes the norm with respect to the covariance matrix $\Sigma = \mathrm{Cov}(a) = \mathrm{Cov}(b)$.

When the transformations are not parameter space perturbations such as dropout, but input space perturbations like translations (which have been found to be far more important to the overall performance of the method \citep{athiwaratkun2018there}), we can show that consistency regularization coincides with minimizing the expected Lie derivative norm. In this sense, consistency regularization can be viewed as an intervention for reducing the equivariance error on unlabeled data.

\subsection{Translation LEE and aliasing}
\label{sec:trans_aliasing_equivariance}
Below we show that spatial aliasing directly introduces translation equivariance error as measured by the Lie derivative, where the aliasing operation $A[\cdot]$ is given by \autoref{eq:alias}.
The Fourier series representation of an image $h(x,y)$ with pixel locations $(x,y)$ 
is $H_{nm}$ with spatial frequencies $(n,m)$, where the band limited reconstruction
\begin{equation*}
    h(x,y) = \tfrac{1}{2\pi}\sum_{nm} H_{nm} e^{2\pi i(xn+ym)} = F^{-1}[H]
\end{equation*}
and $F^{-1}$ is the inverse Fourier transform, and the sums range over frequencies of $-M/2$ to $+M/2$ for both $n$ and $m$ where $M$ is the image height and width (assumed to be square for convenience).

Applying a continuous translation by $t\mathbf{v}$ along vector $\mathbf{v}=(v_x,v_y)$ to the input means resampling the translated band limited continuous reconstruction $h(x,y)$ at the grid points.
\begin{align*}
    T_{t\mathbf{v}}[h](x,y) &= h(x-tv_x,y-tv_y) =\tfrac{1}{2\pi}\sum_{n,m = -M/2}^{M/2}H_{nm} e^{2\pi i [(x-tv_x)n+(y-tv_y)m]}
\end{align*}
To simplify the notation, we will consider translations along only $x$ and suppress the $m$ index of $H_{nm}$, effectively deriving the result for the translations of a $1$d sequence, but that extends straightforwardly to the $2$ dimensional case.
\begin{align*}
    T_{t\mathbf{v}}[h](x) &= h(x-tv_x) =\tfrac{1}{2\pi}\sum_{n = -M/2}^{M/2}[H_{n}e^{-2\pi i t v_xn}]  e^{2\pi i xn}
\end{align*}

Applying the aliasing operation, sampling the image to a new size $M'$ (with Nyquist frequency $M'/2$), we have
\begin{align*}
    A[T_{t\mathbf{v}}[h]](x) &= \tfrac{1}{2\pi}\sum_{n = -M/2}^{M/2}[H_{n}e^{-2\pi itv_xn}] e^{2\pi i x\mathrm{Alias}(n)} \\
    &= \tfrac{1}{2\pi}\sum_{n' = -M'/2}^{M'/2}\bigg[\sum_{n=\mathrm{Alias}^{-1}{(n')}}H_{n}e^{-2\pi itv_xn}\bigg]e^{2\pi i xn'}
\end{align*}
where the last line follows from applying a change of variables $n'=\mathrm{Alias}(n)$.

Applying the final inverse translation (which acts on the $M'$ sampling rate band limited continuous reconstruction), we have
\begin{align*}
    T_{-t\mathbf{v}}[A[T_{t\mathbf{v}}[h]]](x) &=\tfrac{1}{2\pi}\sum_{n' = -M'/2}^{M'/2}\bigg[\sum_{n=\mathrm{Alias}^{-1}{(n')}}H_{n}e^{-2\pi itv_x(n-n')}\bigg]e^{2\pi i xn'}.
\end{align*}
Taking the derivative with respect to $t$, we have
\begin{align*}
    \mathcal{L}_{\mathbf{v}}(A)(h) &= \tfrac{d}{dt}\big|_0 T_{-t\mathbf{v}}[A[T_{t\mathbf{v}}[h]]]\\
    &= \tfrac{1}{2\pi}\sum_{n' = -M'/2}^{M'/2}\bigg[\sum_{n=\mathrm{Alias}^{-1}{(n')}}2\pi iv_x(n'-n)H_{n}\bigg]e^{2\pi i xn'}.
\end{align*}
Notably, for aliasing when the frequency is reduced by a factor of $2$ from downsampling, there are only two values of $n$ that satisfy $\mathrm{Alias}(n)=n'$: the value $n=n'$ and the one that gets aliased down, therefore when multiplied by $n-n'$ the sum 
$$\bigg[\sum_{n=\mathrm{Alias}^{-1}{(n')}}2\pi iv_x(n'-n)H_{n}\bigg]$$ 
consists only of a single term.

According to Parseval's theorem, the Fourier transform $F$ is unitary, and therefore the norm of the function as a vector evaluated at the discrete sampling points $x=1/M',2/M',...$ is the same as as the norm of the Fourier transform:
\begin{align*}
    \|\mathcal{L}_{\mathbf{v}}(A)(h)\|^2 &= \|F[\mathcal{L}_{\mathbf{v}}(A)(h)]\|^2\\
     \|\mathcal{L}_{\mathbf{v}}(A)(h)\|^2 &= \sum_{n'=-M'/2}^{M'/2} \bigg|\sum_{n=\mathrm{Alias}^{-1}{(n')}}2\pi iv_x(n'-n)H_{n}\bigg|^2\\
     \|\mathcal{L}_{\mathbf{v}}(A)(h)\|^2 &= \sum_{n=-M/2}^{M/2} (2\pi)^2 v_x^2(\mathrm{Alias}(n)-n)^2H_{n}^2,
\end{align*}
using the fact that only one element is nonzero in the sum. Finally, generalizing to the $2$d case, we have
\begin{equation}
    \|\mathcal{L}_{\mathbf{v}}(A)(h)\|^2 = (2\pi)^2\sum_{nm}H_{nm}^2\big(v_x^2(\mathrm{Alias}(n)-n)^2+v_y^2(\mathrm{Alias}(m)-m)^2\big),
\end{equation} showing how the translation Lie derivative norm is determined by the higher frequency components which are aliased down.

\section{Learned Equivariance Experiments}

\subsection{Layer-wise Equivariance Baselines}
\label{sec:layerwise-equivariance-baselines}

We use $\text{EQ-T}$ and $\text{EQ-T}_{\text{frac}}$ \citep{karras2021alias} to calculate layer-wise equivariance by caching intermediate representations from the forward pass of the model. For image-shaped intermediate representations, $\text{EQ-T}$ samples integer translations in pixels between -12.5\% and 12.5\% of the image dimensions in pixels. $\text{EQ-T}_{\text{frac}}$ is identical but with continuous translation vectors. The individual layer is applied to the transformed input and then the inverse group action is applied to the output, which is compared with the original cached output. Many different normalization could be chosen to compare equivariance errors across layers. The most obvious are $\frac{1}{N}$, $\frac{1}{\sqrt{N}}$, and $\frac{1}{1}$ (no normalization), where $N = C \times H \times W$. As we show in \autoref{sec:equiv-and-architecture}, the normalization method can have a large effect of the relative contribution of a layer, despite the decision being relatively arbitrary (in contrast to LEE, which removes the need for doing so as the scale is automatically measured relative to the contribution to the output). 

\subsection{Subnetwork Equivariance Analysis}

\begin{figure}[ht!]
    \centering
    \includegraphics[width=0.9\textwidth]{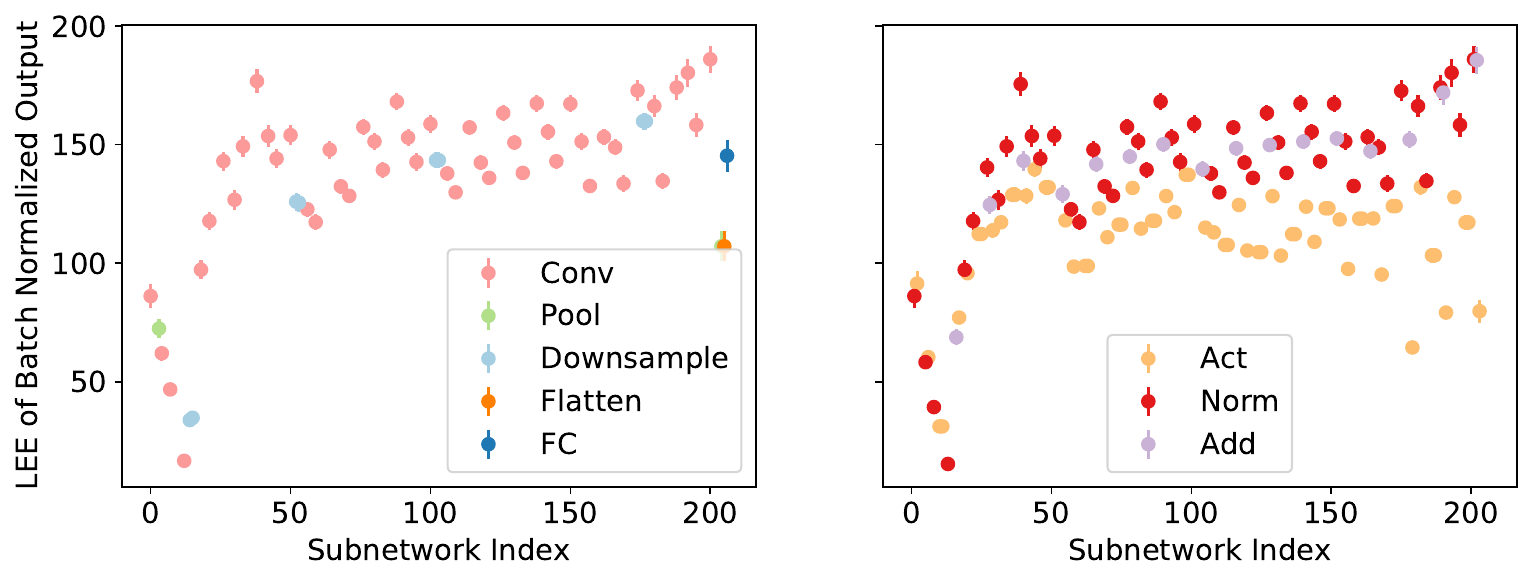} 
    \caption{LEE calculated over the subnetworks of a ResNet50. Specifically a subnetwork is constructed between the input and every intermediate representation in the network's computation graph. We use batch normalization of the outputs to make the output scale of different subnetwork comparable. For visual clarity, layer types are broken across the left and right plots, which share the same axes. Similar to the pattern observed in \autoref{fig:layerwise_architectures}, we see a rapid increase in equivariance error in the early layers of the network, followed by many smaller increases later in the network. Unlike in our layerwise decomposition, comparison across layer types is challenging in this setting because layers have significantly different outputs. For example, comparing activations with preactivations is complicated by the ReLUs acting as contractions of the input and having potentially many zeroed values.}
    \label{fig:subnetworks}
\end{figure}

Another way one might use LEE to study the effects of different layers that make up a network is to break the network in question down into its constituent subnetworks (networks starting at the input and ending at every intermediate representation in the network) and calculate the LEE of the corresponding function. We show the result of this calculation on a ResNet50 in \autoref{fig:subnetworks}. 

As an alternative to our layerwise analysis, this method has a significant drawback that makes analysis challenging: the functions under consideration have different outputs. In our calculation, we applied batch normalization over the outputs in order to make their scales comparable. Despite this rescaling, comparing activations and preactivations, for example, remains challenging. By contrast, our layerwise breakdown specifically targets a layer's contribution to a shared output. 

\subsection{Model List}
\label{sec:model-list}

The models included in Figure 1 are 
\begin{itemize}
    \item Early CNNs: ResNets \citep{he2015deep}, ResNeXts \citep{xie2017aggregated}, VGG \citep{simonyan2014very}, Inception \citep{szegedy2016inceptionv4}, Xception \citep{chollet2017xception}, DenseNet \citep{huang2017densely}, MobileNet \citep{sandler2018mobilenetv2}, Blur-Pool Resnets and Densenets \citep{zhang2019making}, ResNeXt-IG \citep{mahajan2018exploring}, SeResNe*ts \citep{hu2018squeeze}, ResNet-D \citep{he2018bag}, Gluon ResNets \citep{gluoncvnlp2020, zhang2019bag, zhang2020resnest}, SKResNets \citep{li2019selective}, DPNs \citep{chen2017dual}
    \item Modern CNNs: EfficientNet \citep{tan2019efficientnet, tan2021efficientnetv2},  ConvMixer \citep{anonymous2022patches}, RegNets \citep{radosavovic2020designing}, ResNet-RS, \citep{bello2021revisiting}, ResNets with new training recipes \citep{wightman2021resnet}, ResNeSts \citep{zhang2020resnest}, RexNet \citep{han2021rethinking}, Res2Net \citep{gao2019res2net}, RepVGG \citep{ding2021repvgg}, NFNets \citep{brock2021high}, XNect \citep{Mehta_2020}, MixNets \citep{tan2019mixconv}, ResNeXts with SSL pretraining \citep{DBLP:journals/corr/abs-1905-00546}, DLA \citep{yu2019deep}, CSPNets \citep{wang2019cspnet}, ECA NFNets and ResNets \citep{brock2021high},
    HRNet \citep{sun2019highresolution}, MnasNet \citep{tan2019mnasnet}
    \item Vision transformers: ViT \citep{dosovitskiy2020image}, CoaT \citep{dai2021coatnet}, SwinViT \citep{liu2021swin}, \citep{bao2021beit}, CaiT \citep{touvron2021going}, ConViT \citep{d2021convit}, CrossViT \citep{chen2021crossvit}, TwinsViT \citep{chu2021twins}, TnT \citep{han2021transformer}, XCiT \citep{el2021xcit}, PiT \citep{heo2021rethinking}, Nested Transformers \citep{zhang2022nested}
    \item MLP-based architectures: MLPMixer \citep{touvron2021training}, ResMLP \citep{touvron2021resmlp}, gMLP \citep{liu2021pay}, MLP-Mixers with (Si)GLU \citep{rw2019timm}
\end{itemize}

\subsection{Alternative End-to-End Equivariance Metrics}
\label{sec:alt-equivariance-metrics}

\paragraph{Discrete Consistency}

We adopt the consistency metric from \citet{zhang2019making}, which simply measures the fraction of top-1 predictions that match after applying an integer translation to the input (in our case by 10 pixels). Instead of reporting consistency numbers, we report $(1 - \text{\% matching})$, so that consistency because a measure of equivariance error. Equivariant models should exhibit end-to-end invariance, high consistency, and low equivariance error. 

\paragraph{Expected Group Sample Equivariance}
Inspired by work in equivariant architecture design \citep{finzi2020generalizing, hutchinson2021lietransformer}, we provide an additional equivariance metric for comparison against the Lie derivative. Following \citep{hutchinson2021lietransformer}, we sample $k$ group elements in the neighborhood of the identity group element, with sampling distribution $\mathcal{D}(G)$, and calculate the sample equivariance error for model $f$ as $\frac{1}{k}||\rho^{-1}_2(g) f(\rho_1(g)x) - f(x)||$. For translations we take $\mathcal{D}(G)$ to be $\text{Uniform}(-5,5)$ in pixels. 

\paragraph{Versus LEE} There are several reasons why the continuous lie derivative metric is preferable over discrete and group sample metrics. Firstly, it allows us to break down the equivariance error layerwise enabling more fine grained analysis in a way not possible with the discrete analog. Secondly, the metric is less dependent on architectural details like the input resolution of the network. For example, for discrete translations by 1 pixel, these translations have a different meaning depending on the resolution of the input, whereas our lie derivatives are defined as the derivative of translations as a fraction of the input size, which is consistently defined regardless of the resolution. Working with the vector space forming the Lie algebra rather than the group also removes some unnecessary freedom in how one constructs the metric. Rather than having to choose an arbitrary distribution over group elements, if we compute the Lie derivatives for a set of basis vectors of the lie algebra, we have completely characterized the space, and all lie derivatives are simply linear combinations of the computed values. Finally, paying attention to continuous transformations reveals the problems caused by aliasing which are far less apparent when considering discrete transformations, and ultimately the relevant transformations are continuous and we should study them directly.

\subsection{LEE for Additional Transformations}
\label{sec:more-transforms}

Beyond the 3 continuous transformations that we study with Lie derivatives above, there are many more that might reveal important properties of the network. Here we include an three additional transformations--hyperbolic rotation, brightening, and stretch.

\begin{figure}[ht!]
    \centering
    \setlength{\tabcolsep}{0pt}
    \begin{tabular}{cccc}
    \hspace{2mm} Hyperbolic Rot. & \hspace{4mm} Brighten & \hspace{2mm} Stretch & \hspace{4mm} Trace Estimator \\
    \includegraphics[height=0.23\textwidth]{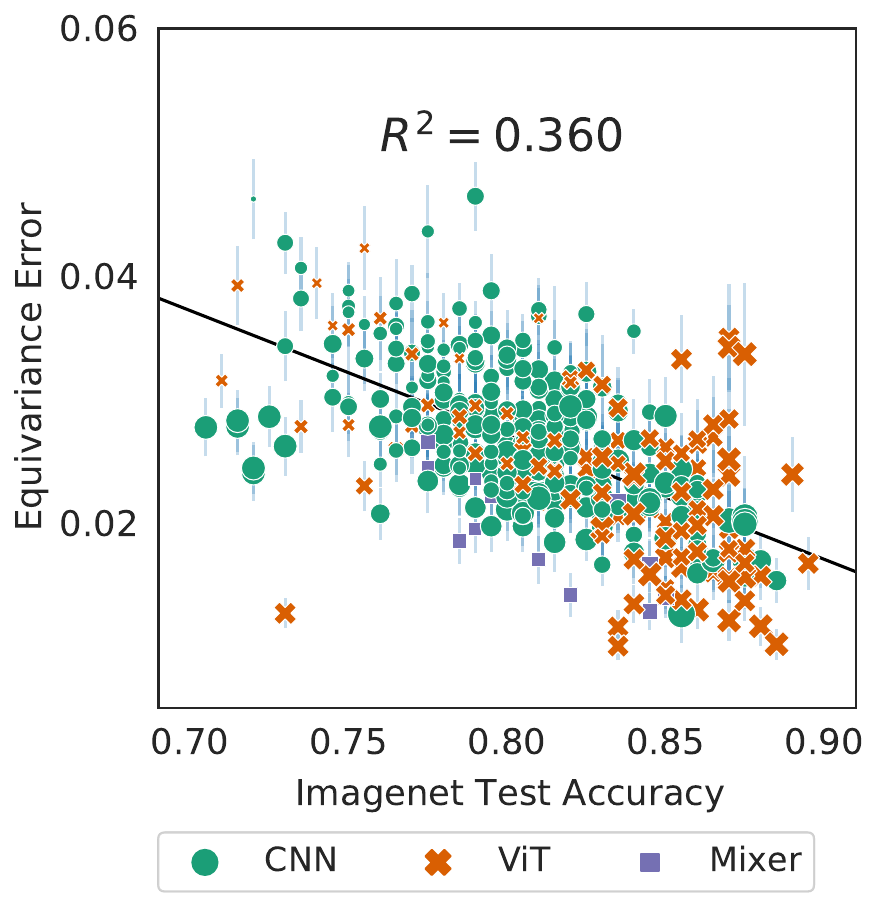} &
    \includegraphics[height=0.23\textwidth]{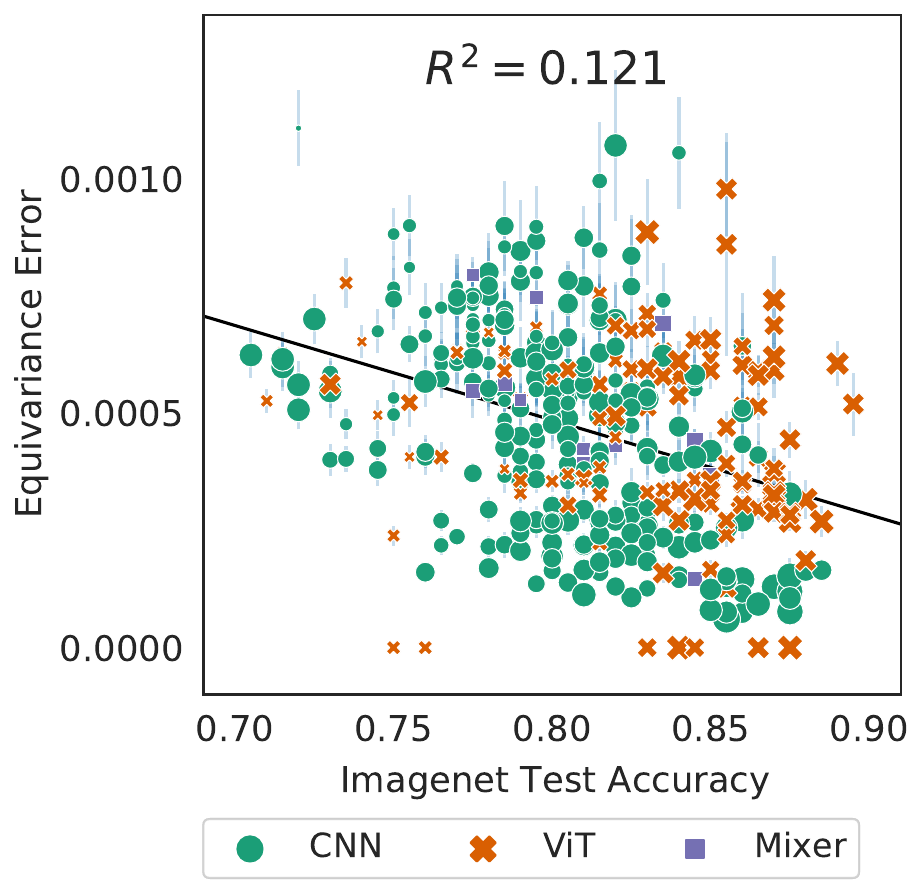} &
    \includegraphics[height=0.23\textwidth]{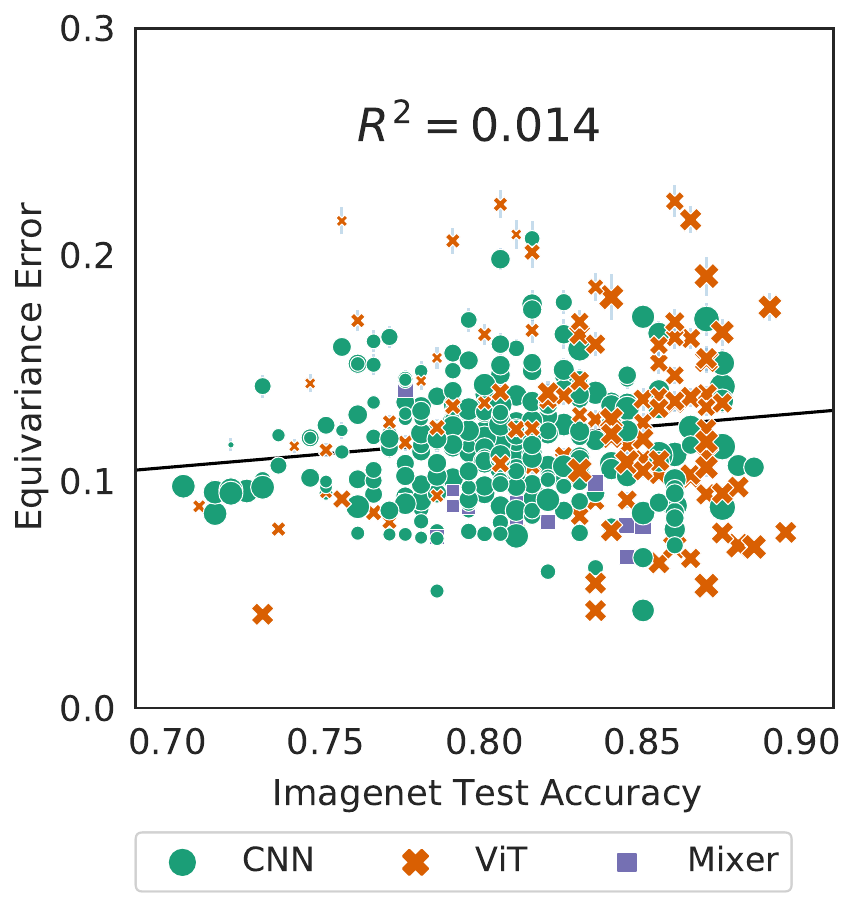} &
    \includegraphics[height=0.23\textwidth]{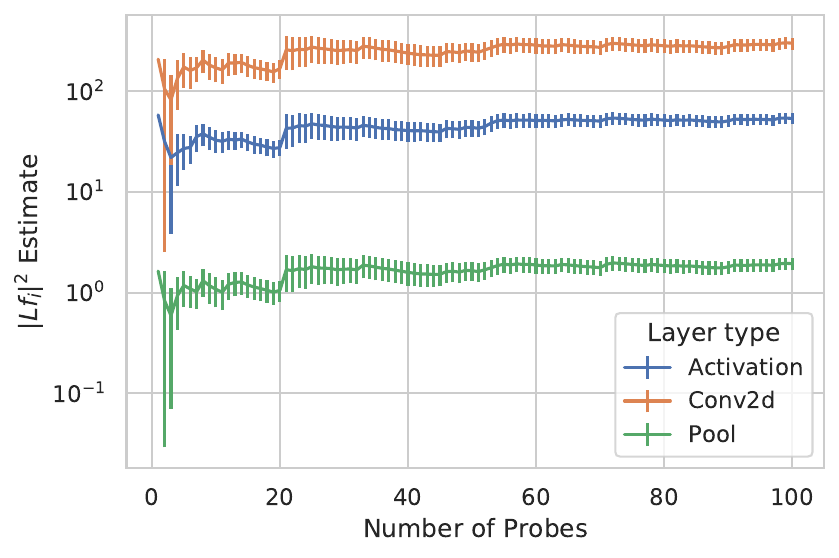}
    \end{tabular}
    \vspace{-5mm}
    \caption{\textbf{(Left)}: Extending \autoref{fig:lie_deriv} we show the Lie derivate norm for  hyperbolic rotation, brightening, and stretch transformations. We observe that more accurate models are also more equivariant to hyperbolic rotations and to brighten transformation, to a more limited extent. In the case of hyperbolic rotations, this result is surprising, as nothing has directly encouraged this equivariance. One possible explanation is decreased aliasing in models with higher accuracy. Marker size indicates model size. Error bars show one standard error over the images use to evaluate the Lie derivative. \textbf{(Right)}: Cumulative mean and standard error of the estimator (computed for translations on a ResNet-50).}
    \label{fig:shear-hyperbolic}
\end{figure}

\autoref{fig:shear-hyperbolic} (left) shows that, perhaps surprisingly, models with high accuracy become more equivariant to hyperbolic rotations. We suspect this surprisingly general equivariance to diverse set of continuous transformations is probably the result of improved anti-aliasing learned implicitly by more accurate models. LEE does not identify any significant correlation between brightening or stretch transformations and generalization ability. 

\subsection{Rotated MNIST Finetuning}
\label{sec:mae-finetune}

In order to test the ability of SOTA imagenet pre-trained models to learn equivariance competitive with specialized architectures, we adapted the example rotated MNIST \href{https://github.com/QUVA-Lab/e2cnn/blob/master/examples/model.ipynb}{notebook} available in E2CNN repository \citep{e2cnn}. We use the \href{https://dl.fbaipublicfiles.com/mae/pretrain/mae_pretrain_vit_base.pth}{base model} and default \href{https://github.com/facebookresearch/mae/blob/main/FINETUNE.md}{finetuning procedure} from \citep{MaskedAutoencoders2021}, finetuning for 100 epochs, halving the learning rate on loss plateaus. 

\end{document}

%% file: math_commands.tex
\usepackage{amsmath,amsfonts,bm}
\usepackage[utf8]{inputenc} 
\usepackage[T1]{fontenc}    
\usepackage{hyperref}       
\usepackage{url}            
\usepackage{amsfonts}       
\usepackage{nicefrac}       
\usepackage{graphicx}
\usepackage{placeins}
\usepackage{subfigure}
\usepackage{dsfont}
\usepackage{xspace}
\usepackage{amssymb}
\usepackage{mathtools}
\usepackage{hyperref}
\usepackage{amsfonts}
\usepackage{color}

\usepackage{float}
\usepackage{scalerel,stackengine}
\usepackage{soul}
\usepackage[algo2e,ruled,vlined]{algorithm2e}
\include{pythonlisting}
\usepackage{appendix}
\usepackage{amsthm}

\usepackage{wasysym}
\usepackage{pifont}
\usepackage{bbm}
\usepackage{lipsum}

\def\eqref#1{equation~\ref{#1}}

\def\1{\bm{1}}

\def\vb{{\bm{b}}}

\def\vx{{\bm{x}}}

\DeclareMathAlphabet{\mathsfit}{\encodingdefault}{\sfdefault}{m}{sl}
\SetMathAlphabet{\mathsfit}{bold}{\encodingdefault}{\sfdefault}{bx}{n}

\newcommand{\R}{\mathbb{R}}

